# Survey of Explainable Machine Learning with Visual and Granular Methods beyond Quasi-explanations

Boris Kovalerchuk[1], Muhammad Aurangzeb Ahmad[2,3], Ankur Teredesai[2,3]

[1]Dept. of Computer Science, Central Washington University, USA
`borisk@cwu.edu`
[2] Dept. of Computer Science and Systems, University of Washington Tacoma, USA
[3]Kensci Inc., USA
`maahmad@uw.edu, ankurt@uw.edu`

**Abstract**. This chapter surveys and analyses visual methods of explainability of Machine Learning (ML) approaches with focus on moving from quasi-explanations that dominate in ML to actual domain-specific explanation supported by granular visuals. The importance of visual and granular methods to increase the interpretability and validity of the ML model has grown in recent years. Visuals have an appeal to human perception, which other methods do not. ML interpretation is fundamentally a human activity, not a machine activity. Thus, visual methods are more readily interpretable. Visual granularity is a natural way for efficient ML explanation. Understanding complex causal reasoning can be beyond human abilities without "downgrading" it to human perceptual and cognitive limits. The visual exploration of multidimensional data at different levels of granularity for knowledge discovery is a long-standing research focus. While multiple efficient methods for visual representation of high-dimensional data exist, the loss of interpretable information, occlusion, and clutter continue to be a challenge, which lead to quasi-explanations. This chapter starts with the motivation and the definitions of different forms of explainability and how these concepts and information granularity can integrate in ML. The chapter focuses on a clear distinction between quasi-explanations and actual domain specific explanations, as well as between potentially explainable and an actually explained ML model that are critically important for the further progress of the ML explainability domain. We discuss foundations of interpretability, overview visual interpretability and present several types of methods to visualize the ML models. Next, we present methods of visual discovery of ML models, with the focus on interpretable models, based on the recently introduced concept of General Line Coordinates (GLC). This family of methods take the critical step of creating visual explanations that are not merely quasi-explanations but are also domain specific visual explanations while these methods themselves are domain-agnostic. The chapter includes results on theoretical limits to preserve n-D distances in lower dimensions, based on the Johnson-Lindenstrauss lemma, point-to-point and point-to-graph GLC approaches, and real-world case studies. The chapter also covers traditional visual methods for understanding multiple ML models, which include deep learning and time series models. We illustrate that many of these methods are quasi-explanations and need further enhancement to become actual domain specific explanations. The chapter concludes with outlining open problems and current research frontiers.

# 1. Introduction

## 1.1. What are explainable and explained?

The terms used to make sense of machine learning models and their predictions has evolved and grown over time. These terms include *interpretability, explainability, comprehensibility, intelligibility*, and *understandability* that often are used interchangeably.

It is important to draw a line between them, and *actual interpretation, explanation, comprehension, and understanding* of ML models, and their predictions. While this difference seems subtle, it is very important. The first category of terms points to the **ability/opportunity** to explain a model, but not necessary getting an **actual** explanation. Consider, a branch of a decision tree or a logic model for a case $\mathbf{x} = (x_2, x_2, x_3)$:

If $(x_1 >5)$ & $(x_2 <7)$ & $(x_3 >10)$ then $\mathbf{x}$ belongs to class 1.

This model can be quite accurate on training, validation and independent test data. A domain expert understands what this model says if attributes $x_1$-$x_3$ are *meaningful* in the domain where the data are taken from. Therefore, this model is *interpretable/understandable* for a *domain expert,* but *not necessary explained* for this expert.

The domain expert can say that despite its high empirical confirmation, it is not clear *why this model should work*. The model *is not explained in the terms of the domain knowledge such as causal relations known in the domain*. This is a quite common situation in ML [Kovalerchuk, et al., 2001], and science in general, when empirical discoveries precede a domain theory on the subject, e.g., empirical Kepler law in physics. Thus, *explained* and explainable models fundamentally *differ*.

The mismatch between ML model and domain knowledge can be of multiple types including concepts, terminology, different levels of granularity, and modality (e.g., free text, tables, formulas, visuals). For granularity difference, a doctor is sure that *high blood pressure* is relevant to the heart problems, but much less sure about the numeric value of a fuzzy term "high".

In [Kovalerchuk, et al., 2001] this issue of mismatch was resolved by building two models: *expert-based models* and *data-based model*s. The **expert-based model** is a *qualitative model,* where the radiologist produces expert classification rules using linguistic terms such as "high" and "large". In contrast the **data-based model** uses a threshold between "large" and "small" to produce binary attributes and logical rules/models, then consistency between two types of models is analyzed by the expert to confirm or reject the data-based model on the ground of domain interpretability.

A model is **explained** if a domain expert accepts it based on both
- *empirical evidence of enough accuracy* and
- the *domain knowledge*/theory/reasoning, which is beyond a given dataset.

A model is **explainable** if a domain expert
- *understands* what the model says,
- understand how to *apply* it to new cases, but
- does *not* accept/view it yet as *consistent* or causal model, based on domain knowledge/theory/reasoning, which is beyond a given dataset.

Sometimes the important, but subtle difference between actual explanation and explainability is presented as follows: "…how well a human *could* understand the decisions in the given context, which is often called interpretability or explainability; and (2) *explicitly* explaining decisions to people, which we will call e*xplanation*" [Miller, 2019]. We made keywords italic in this quote to emphasize the difference. Thus, an *explainable* model can be very far from a needed *explained domain-relevant,* or a *causal model*.

In other words, the model is **explainable** if
- it is presented *only* in the *domain terms* (e.g., medicine) without terms that have no meaning in the domain yet.

Note, that many ML algorithms often derive the predictive models and their explanations using ML terms that are *foreign* to the domain, like distances, weighs, hidden layers and so on. We describe such non-domain explanations as "quasi-explanations" since they do not make domain sense. These explanations can make sense for the data scientists, but it is not surprising that domain experts call these models "black-boxes", and tend to avoid using them.

Respectively, we will call a model **quasi-explainable** if
- its explanation statements contain terms that are *foreign* for the domain.

One of the major goals of this chapter is showing ways toward deep explanation, which minimize or avoid quasi-explanations. We are not calling to stop development of ML models with only quasi-explanations, but merely highlighting that quasi-explanations, are not the final goal. The visual knowledge discovery approach based on General Line Coordinates (GLC), presented in section 5, does not use terms, which are foreign to the domain, in models that it constructs, and its explanations are domain specific visual explanations.

## 1.2. Types of machine learning models

The major classes of ML models are **black box** models and **glass box** or **white box** models. The former does not explain how the model is making predictions, or provides a very limited quasi-explanation. These models keep a user in dark on explanation. They require significant extra effort for an explanation if it is possible. Examples of such models are SVMs, random forest, CNN etc. Multiple deficiencies of black-box machine learning models are discussed in the literature (e.g. [Liao et al, 2020; Liu, 2019; Rudin, 2019; Zhang et al, 2020]

The glass box models explain how they predict, or an explanation can be added quite directly. Examples are decision trees, logical rules etc. A popular statement is that black box models are less interpretable, but more accurate, than glass box models, while alternative statements are getting momentum [Rudin, 2019; Neuhaus, Kovalerchuk, 2019] stating that glass boxes can achieve both. The reason for this is to avoid choosing between accuracy and interpretability, which is a major obstacle to the wider adoption of Machine Learning in areas with high cost of error, such as in cancer diagnosis, and many other domains, where it is necessary to understand, validate, and trust decisions. Visual knowledge discovery methods based on General Line Coordinates (GLC) [Kovalerchuk, 2018], surveyed in this chapter, help to get both the model accuracy and its explanation beyond the quasi-explanations.

**Quasi-explainable linear models**. Linear regression and discrimination models often are listed as interpretable. "The linearity of the learned relationship makes the interpretation easy" [Molnar, 2020]. In general, this is an **incorrect statement**. Linear regression and discrimination models in ML are multidimensional models. They only can be relatively easily interpretable if all attributes are **homogeneous**. e.g., measure the same single attribute, such as temperature, or stock price at different moments in *time series* prediction. However, typically in ML, attributes are **heterogeneous** as in medical diagnostics: blood pressure, cholesterol level, temperature and so on in a single dataset for a given time. The weighted summation of such heterogeneous attributes does not have much physical meaning. For instance, the sum of the blood pressure and temperature is not a part of the medical language, such sum is not defined meaningfully. Thus, an explanation that uses *summation of heterogeneous attributes* can be at best a **quasi-explanation**, but definitely **not a deep explanation**.

Therefore, we need to be very careful claiming that regression models are interpretable. Even when attributes are homogeneous it's still *not necessary* that the regression models will be *meaningful*. For instance, what is the meaning of a weighted sum of systolic and diastolic blood pressure measurements? While both measure blood pressure inside the arteries, systolic one measures it when the heart is pumping, but the diastolic one measures it when the heart is resting between beats. Additionally, a linear regression model as well as a deep learning model may be using highly engineered features e.g., summation of cube root of bodily vital stats which may not have a domain interpretation.

**How to interpret a linear model?** We can solve the problem of interpretation of a linear model, by interpolating it by a set of *logical rules*. In this way, a quasi-explainable model is converted into an explainable one.

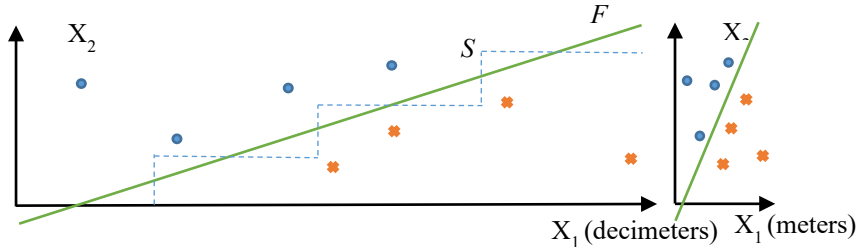

Fig. 1. Linear discrimination function $F(x,y)$ interpolated by a step function $S$ with attribute $X_1$ measured in decimeters (left) and meters (right)

The idea of this interpolation is illustrated in Fig. 1. It shows a green discrimination line $F(x_1,x_2) = ax_1+bx_2+c = 0$. Its discrimination rule is: If $F(x,y) > 0$, then $(x,y)$ belongs to class 1 else to class 2. Next the green line is interpolated by a dotted *step function S*. Then each step $i$ is represented as a logical rule:

If $x_1$ is in the limits $s_{i11} \leq x_1 \leq s_{i12}$, and $x_2$ is in limits $s_{i21} \leq x_2 \leq s_{i22}$ defined by the step, then pair $(x_1, x_2)$ is in class 1, else in class 2.

We interpolate a whole linear function with multiple steps, and multiple such logical rules. This set of rules can be quite large in contrast with a compact linear equation, but the rules have a clear interpretation for the domain experts when $x_1$ and $x_2$ have a clear meaning in the domain.

For each new case to be predicted, we just need to find a specific step, which is applicable to this case and provide a single **simple local rule** of that step to the user as an explanation. Moreover, we do not need to store the step function and rules. They can be generated on the fly for each new case to be predicted.

To check that the linear function is meaningful before applying to new cases, a user can randomly select validation cases, find their expected steps/rules and evaluate how meaningful those rules and their predictions are.

**Quasi-explainable weights**. Often it is claimed that weights in the linear models are major and efficient tool to provide model interpretation to the user, e.g., [Molnar, 2020]. Unfortunately, in general this is an **incorrect statement,** while multiple AutoML systems implement it as a model interpretation tool [Xanthopoulos et al, 2020] for linear and non-linear discrimination models. The reason is the same as before – heterogeneity of the attributes. If a weighted sum has no interpretation, then the fact the weight $a$ for $x_1$ is two time greater than $b$ for $x_2$ has no meaning to be able to say that $x_1$ is two times more important than $x_2$. It is illustrated in Fig 1.

In the case of homogeneous attributes and a vertical discrimination line, a single attribute $X_1$ would discriminate the classes. Similarly, if a horizontal line discriminates classes then a single attribute $X_2$ would discriminate classes, and $X_1$ would be unimportant. If the discrimination line is the diagonal with 45°, then $X_1$ and $X_2$ are seem equally important. Respectively, for the angles larger than 45° $X_1$ would be more important else $X_2$ would be more important.

In fact, in Fig. 1 (on the left) the angle is less than 45° with $X_2$ more important, but on the right the angle is greater than 45° with $X_1$ more important. However, both pictures show the discrimination for the same data. The only difference is that $X_1$ is in decimeters on the left and in meters on the right. If $X_1$ is a length of the object in decimeters and $X_2$ is its weight in kilograms, then $X_2$ is more important in Fig.1, but if $X_1$ is in meters, then $X_1$ is more important. Thus, we are getting very different relative importance of these attributes. In both cases we get *quasi-explanation*. In contrast if $X_1$ and $X_2$ would be homogeneous attributes, e.g., both lengths in meters, then weights of attributes can express the importance of the attributes meaningfully and contribute to the actual not quasi-explanation.

Note, that a linear model on heterogeneous data converted to logical rules is free of this confusion. The step intervals are expressed in actual measurement units.

The length and height of those steps can help to derive the importance of attributes. Consider a narrow, but tall step. At first glance it indicates high sensitivity to $X_1$ and low sensitivity to $X_2$. However, it depends on meaningful insensitivity units in $X_1$ and $X_2$, which a domain expert can set up. For instance, let the step length be 10 m, the step height be 50 kg with insensitive units of 2 m and 5 kg. Respectively, we get 5 $X_1$ units and 10 $X_2$ units and can claim high sensitivity to $X_1$ and low sensitivity to $X_2$. In contrast, if the units will be 2 m and 25 kg, then we get 5 $X_1$ units and 2 $X_2$ units then we can claim the opposite. Thus, a meaningful scaling and insensitivity units needs to be set up as part of the model discovery and interpretation process for linear models with heterogeneous attributes to avoid quasi-explanations.

## 1.3. Informal definitions

Often desirable characteristics of the explanation are used to define it, e.g., [Craik 1952, Doshi-Velez, Kim, 2017; Doran et al. 2017]:

- Give an explanation that in meaningful to a **domain expert**;
- Give an explanation comprehensible to humans in (i) **natural language** and in (ii) **easy** to understand **representations**;
- Give an explanation to **humans** using **domain knowledge** not ML concepts that are **external** to the domain;
- Provide **positive** or **negative** arguments for the prediction;
- Answer why is this prediction being made not an **alternative** one?
- Give an explanation how inputs are mathematically **mapped** to outputs e.g., regression and generalized additive models.

The question is how to check that these desirable characteristics are satisfied. Microsoft and Google started eXplainable AI (XAI) services, but do we have **operational definitions** of model comprehensibility, interpretability, intelligibility, explainability, understandability to check these properties? Is a bounding box around the face with facial landmarks provided by the Google service an operational explanation without telling in understandable terms how it was derived?

## 1.4. Formal operational definitions

It is stated in [Molnar, 2020]: "There is no mathematical definition of interpretability." Fortunately, this statement is not true. Some definitions are known for a long time. They require

- showing how each training example can be inferred from background knowledge (domain theory) as an instance of the target concept/class by *probabilistic first order logic* (FOL), e.g., [Mitchell, 1997; Muggleton, 1992, 1996; Kovalerchuk, Vityaev, 2000].

Below we summarize more recent ideas from [Muggleton et al, 2018], which are inspired by work of Donald Michie [1988], who formulated three Machine Learning quality criteria:

- Weak criterion – ML improves predictive *accuracy* with more data.
- Strong criterion – ML additionally provides its *hypotheses* in symbolic form.
- Ultra-strong criterion (**comprehensibility**) – ML additionally *teaches the hypothesis to a human*, who consequently performs *better* than the human studying the training data alone.

The definitions from [Muggleton et al, 2018], presented below, are intended to be able to test the ultra-strong criterion. These definitions allow studying comprehensibility experimentally and operationally.

**Definition** (*Comprehensibility, C(S, P)*)
The comprehensibility of a definition (or program) $P$, with respect to a human population S, is the mean accuracy with which a human s from population S after brief study and without further sight, can use $P$ to classify new material, sampled randomly from the definition's domain

**Definition** (*Inspection time T (S, P)*)

The inspection time T of a definition (or program) *P*, with respect to a human population S, is the mean time a human s from S spends studying P, before applying P to new material.

**Definition** (*Textual complexity, Sz(P)*)

The textual complexity Sz, of a definition of definite program *P*, is the sum of the occurrences of predicate symbols, functions symbols and variables found in *P*.

The ideas of these definitions jointly with prior *Probabilistic First Order Logic* inference create a solid mathematical basis for interpretability developments.

### 1.5. Interpretability and granularity

It was pointed out in [Choo et al, 2018] that *interpretation* differs from *causal reasoning* stating that interpretation is much more than causal reasoning. The main point in [Choo et al, 2018] is that, even if we will have perfect causal reasoning for the ML model, it will not be an explanation for a human. A detailed causal reasoning can be *above human abilities* to understand it. Assume that this is true, then we need **granularity of the reasoning,** generalized at different levels, in linguistic terms. It can be similar to what is done in fuzzy control. In fuzzy control, a human expert formulates simple control rules, in uncertain linguistic terms, without details such as if A is *large* and B is *small,* and C is *medium,* then control D needs to be *slow*.

In fuzzy control, such rules are formalized through membership functions (MFs) and different aggregation algorithms, which combine MFs. Then the parameters of MFs are tuned using available data. Finally, the produced control model became a competitive one with a model built using solid physics, as multiple studies had shown.

The ML explanation task of detailed causal reasoning is an *opposite task*. We do not start from a linguistic rule, but from a complex reasoning sequence and need to create simple rules. Moreover, these rules need to be at different levels of detail/granularity. How can we design this type of rules, from black box machine learning models?

This is a very complex and open question. First, we want to build this type of granular rules, for the explainable models, such as decision trees. Assume that we have a huge decision tree, with literally hundreds of nodes, and branches with dozens or hundreds of elements on each branch. A human can trace and understand any small part of it. However, the total tree is beyond the human capabilities for understanding and tracing a branch which has, say, 50 different conditions like $x_1 > 5$, $x_2 < 6$, $x_3 > 10$ and so on 50 times. It is hard to imagine that anybody will be able to meaningfully trace and analyze it easily. In this situation, we need ways to *generalize* a decision tree branch, as well as the whole tree. Selecting most important attributes, grouping attributes to larger categories, and matching with a representative case are among the ways to decrease human cognitive load. Actual inequalities like $x_1 > 5$, $x_2 < 6$, and $x_3 > 10$ can be substituted by "large" $x_1$ and $x_3$, and "small" $x_2$. If $x_1$ and $x_3$ are width and length and $x_2$ is weight of the object, then we can say that a large light object belongs to class 1. If an example of such an object is a bucket, we can say that objects, like buckets, belong to class 1. The perceptually acceptable visual explanation of the ML is often at the different and more coarse level of granularity than the ML model. Therefore, visual ML models that we discuss in section 4, have important advantage being at *the same granularity level* as their visual explanation.

## 2. Foundations of Interpretability

### 2.1. How interpretable are the current interpretable models?

Until recently the most interpretable large time series models were not really interpretable [Schlegel et al., 2019]. In general, the answer for this question depends on many factors, such as the definition of the interpretable models and the domain needs of explanation. We already discussed the definition issue.

The needs issues are as follows: How severe the domain needs the explanation, what kind of explanation would suffice the domain needs, how complete an explanation is needed, and types/modality of data. These factors are summarized below:

- *Level of domain needs*. Problems in healthcare e.g., risk of mortality have more stringent explanation requirement than retail e.g., placement of ads [Ahmad 2018].
- *Level of soundness needed.* An explanation is sound if it adheres to how the model works [Kulesza et al, 2015]. The optical character recognitions (OCR) model for the text printed in a high-quality laser printer based on the neural networks does not need to be sound as far as it has high OCR accuracy. In contrast, models for healthcare need to be sound.
- *Level of completeness needed*. An explanation is complete if it encompasses the complete extent of the model [Kulesza et al, 2015]. The OCR task of the handwritten text needs explanation for poorly written characters and people with bad handwriting habits.
- *Data types/modality*. Different data types (structured or unstructured data, images, speech, text, time series and so on) can have different needs for explanation.

### 2.2. Domain specificity of interpretations

Domain specificity has multiple aspects. The major one is the need to describe the trained ML model in terms of **domain ontology**, without using terms that are *foreign* to the domain, where the ML task must be solved [Kovalerchuk, 2020]. It is much more critical for the domains and problems with high cost of errors, such as medicine.

The next question is: to what extent the terms and concepts, which are foreign for the domain can be included into the explanations, and continue to be useful. The general statements like the explanations/interpretation must make sense to the domain expert, who is going to use the ML model, are not very helpful because they are not operational.

Another important question is what should be the explanation for the *domain without much theory and background knowledge*? An example is predicting new movie rating. The same question should be answered for the domain, where the background knowledge is very *inconsistent*, and expert opinions are very diverse on the same issue.

### 2.3. User centricity of interpretations

While asking the explanations to be in the right language, and in the right context [Doshi-Velez et al, 2017, Druzdzel 1996] seems mandatory, the actual issue is how to define it within the domain terms and ontology. Not every term from the domain ontology can be used in the explanation efficiently. Moreover, one of the equivalent concepts can be preferred by some users. Some users can prefer decision trees, while others prefer logic rules when both are applicable.

The quest for simple explanation came to its purest form in the ELI5 principle: Explain it Like I am 5 years old. It is obviously not coming for free, making sense may require sacrificing or deemphasizing model fidelity. User-centricity of the explanations often requires to be **role-based**. I physician needs different explanations as compared to a staffing planner in a hospital. Thus, simplicity of the explanation for people in these roles is not the same.

### 2.4. Types of Interpretable models

**Internally interpreted vs. externally interpreted models**. *Internally interpreted models* do not separate the predictive model and the explanation model. Such models are *self-explanatory*, e.g., decision trees. They are explained in terms of interpreted elements of their structure not only inputs. An extra effort to convert the model to a more convenient form, including visualization, often is beneficial.

*Externally interpreted models* contain a separate predictive model and an explanation model and an external explanation model. Such ML models are explained in terms of interpretable input data and attributes, but typically without interpreting the model structure. An example would be providing a list of most important input attributes without telling how it is supported by the model's structure. Another example is producing an interpretable decision tree or logic rules from the neural networks (NN) [Shavlik, 1996] by interpolating a set of input-output pairs generated from the NN. As any interpolation it can differ from the NN, for instance, due to the insufficient size of the set. If a new case to be predicted is not represented in the generated set of input-output pairs by similar cases, then the decision tree prediction and explanation will not represent the NN model. This is a major problem of the external explanation approach that can produce a quasi-explanation.

**Explicit vs. Implicit Interpretations.** Decision trees and logic rules provide explicit interpretations. Many other ML methods produce only implicit interpretations where a full interpretation needs to be derived using additional domain knowledge. An implicit heatmap explanation for a Deep Neural Network (DNN) model, which we discuss in detail in the next section, requires human knowledge beyond the image.

### 2.5. Using black-box models to explain black box models

The examples below show **shallow not deep explanations** that attempt explaining **one black box using another black box**. Often the end users very quickly recognize this because black box explanation does not answer a simple question: Why is this the right explanation? Deciphering that black box is left for the user.

Consider a task of recognizing a boat in Fig. 2 from [Montavon et al., 2018] with an implicit DNN explanation. We can recognize a boat based on a group of pixels highlighted by DNN as a heatmap that represent a mast.

To derive a *conceptual explanation* that uses the *concept* of a mast we need external human common-sense knowledge *what is the mast* and how it differs from other objects. This is not a part of the heatmap DNN explanatory model that shows salient pixels. Without the concept of the mast salient pixels provide *a shallow black box explanation not a deep explanation*. Thus, **deep learning neural network** models, are **deep in terms of the number of layers** in the network. This number is larger than in the prior traditional neural network models. However, as was illustrated above, DNNs are **not deep in the terms of the explanations**. Fundamentally, new approaches are needed to make the current quasi-explanation in DNN a really deep explanation. An alternative approach, which we advocate, such as GLC, is building explainable models from the very beginning, in addition or instead of explaining DNN, and other black boxes.

Similarly, in medical imaging, external domain knowledge is needed for **deep explanation**. If an expert radiologist cannot match DNN salient pixels with the domain concepts such as tumor, these pixels will not serve as an explanation for the radiologist. Moreover, the radiologist can reject these pixels to be a tumor.

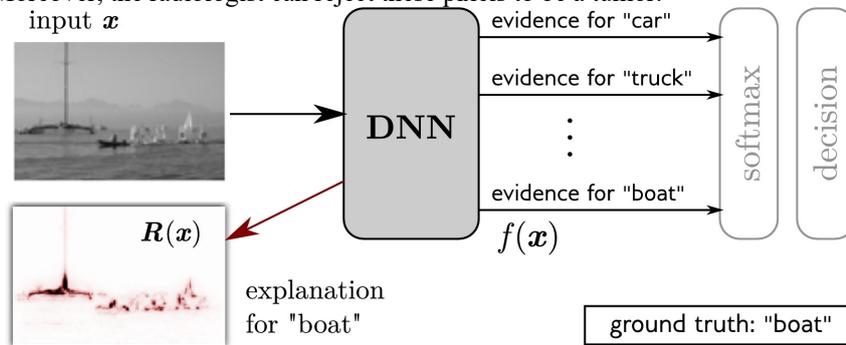

Fig. 2. DNN boat explanation example [Montavon et al, 2018]

The major problem is explaining, in the **domain terms, why** are these salient points are right ones. It fundamentally differs from explaining, in **ML terms, how** these points were derived. One of the common methods, for the last one, *is backward gradient tracing* in DNN, to find salient pixels that contributed most to the class prediction. This explanation is completely *foreign* to the radiology domain. In other words, we try to produce an explanation, using unexplained and unexplainable method for the domain expert. This can be a deep explanation for the computer scientist, not for a radiologist who is the end user.

In the boat example, in addition to the unexplained prediction of class "boat", we produce unexplained salient pixels as an explanation of the boat. Here we attempt to explain **one black box**, **using another black box**. This is a rather **quasi-explanation**. This quasi-explanation happened, because of the use of model concepts and structures, which are foreign to the domain.

**Why explanation models are often not explained?** Often it is just a reflection of the fact, that the ML model explainability domain, is in the *nascent stage*. Explaining one black box, using another black box, is an acceptable first step to deep explanation, but it should not be the last one. Can every black box explanation be expanded to a deep one? This is an open question.

## 3. Overview of Visual Interpretability

### 3.1. What is visual interpretability?

Visual methods that support interpretability of ML models have several important advantages, over non-visual methods, including faster and more attractive communication to the user. There are four types of visual interpretability approaches, for ML models, and workflow processes of discovering them:
(1) *Visualizing existing ML models* to support their interpretation;
(2) *Visualizing existing workflow processes* of discovering ML models to support their interpretation;
(3) **Discovering new interpretations of existing ML models**, and processes by using visual means;
(4) **Discovering new interpretable ML models** by using visual means.

The goal of (1) and (2) is *better communicating* on existing models and processes, but not discovering new ones using visual means. Visualization of salient points with heatmap in DNN is an example of (1). Other works exemplify (2): they visualize *specific points* within the workflow process (hyperparameter tuning, model selection , the relationships between a model's hyperparameters and performance), provide multi-granular visualization, and monitor the process and adjust the search space in real time [Vizier [Golovin et al, 2017; Park, et al; Wang et al, 2019]. In contrast, AutoAIVis system [Weidele et al, 2020] focuses on multilevel real-time visualization, of the *entire process*, from data ingestion to model evaluation using Conditional Parallel Coordinates [Weidele, 2019].

The types (3) and (4) are potentially much more rewarding, but more challenging, while many current works focus on (1) and (2). This chapter focuses on types (2) and (4). The last one can produce interpretable models, avoiding a separate process of model interpretation.

### 3.2. Visual vs. non-visual methods for interpretability and why visual thinking

Fig. 3 illustrates benefits of visual understanding over non-visual ones. Analysis of images is a parallel process, but analysis of the text (formulas and algorithms) is a sequential process, which can be insufficient.

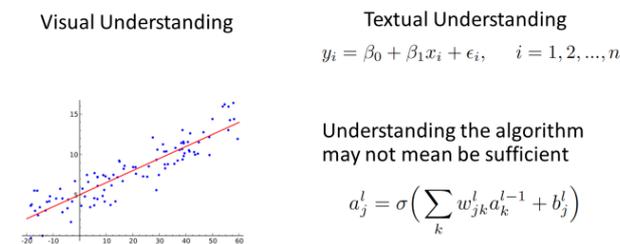

Fig. 3. Visual understanding vs. over non-visual understanding.

Chinese and Indians knew a visual proof of the Pythagorean Theorem in 600 B.C. before it was known to the Greeks [Kulpa, 1994]. Fig. 4 on the left shows it. This picture was accompanied by a single word see as a textual "explanation" with it.

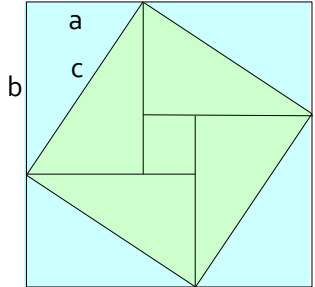

Fig. 4. Ancient proof (explanation) of the Pythagorean Theorem. Actual explanation of the theorem presented in a visual form.

To provide a complete analytical proof the following inference can be added in modern terms: $(a+b)^2$ (area of the largest square) - $2ab$(area of 4 blue triangles) = $a^2+b^2$ = $c^2$ (area of inner green square).

Thus, we follow this tradition -- moving **from visualization of solution to finding a solution visually** with modern data science tools. More on historical visual knowledge discovery can be found in [Kovalerchuk, Schwing, 2005].

### 3.3. Visual interpretation pre-dates formal interpretation

Fig. 5 shows an example of visual model discovery in 2-D, for the data in the table on the left [Kovalerchuk, 2020]. Here, a single fitted black line cannot discriminate these two "crossing" classes. In addition, the visualization clearly shows, that any single line cannot discriminate these classes.

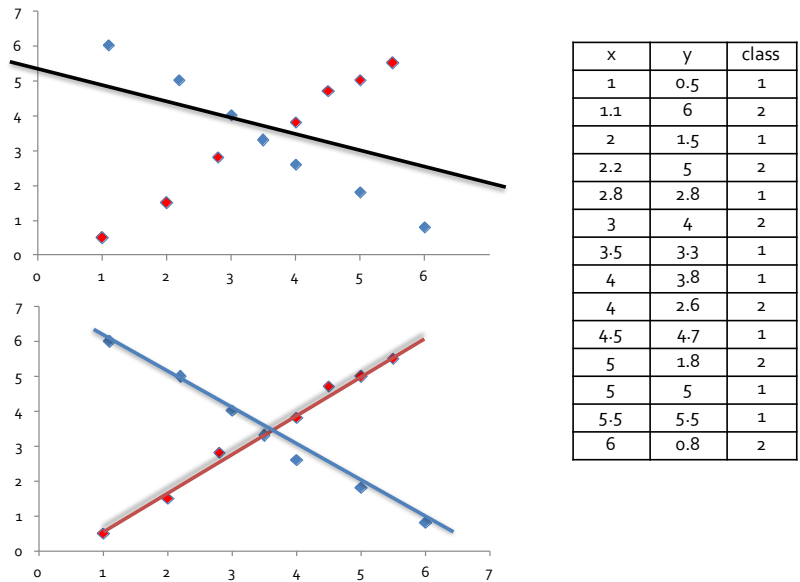

Fig. 5. "Crossing" classes that cannot be discriminated by a single straight line.

| ID | FD1 | FD2 | FD4 | FD5 | FD6 | FD10 | FD12 | FD15 | FD16 | FD18 | FD20 | FD22 | FD23 | FD24 | FD25 | FD26 | FD27 | FD28 |
|---|---|---|---|---|---|---|---|---|---|---|---|---|---|---|---|---|---|---|
| 1 | 0 | 0 | 2.749807 | 9.826302 | 4.067554 | 0 | 0 | 0 | 5.244006 | 0 | 2.743422 | 0 | 0 | 0 | 0 | 6.254963 | 0 | 0 |
| 2 | 11.51334 | 9.092989 | 0 | 12.46223 | 0 | 7.597155 | 0 | 0 | 8.940897 | 0 | 0 | 0 | 4.268456 | 0 | 0 | 0 | 0 | 1.309903 |
| 3 | 10.27931 | 0 | 2.075787 | 0 | 0 | 4.042145 | 0 | 0.477713 | 3.97378 | 0 | 0 | 2.477745 | 0 | 0 | 0 | 5.583099 | 0 | 7.418219 |
| 4 | 0 | 18.31495 | 0 | 0 | 0 | 0 | 0 | 0 | 4.472742 | 4.671682 | 0 | 7.248355 | 12.11645 | 0 | 0 | 0 | 6.030322 | 0 |
| 5 | 14.12261 | 15.1236 | 9.695051 | 0 | 0.915031 | 0 | 0 | 0 | 6.086389 | 9.139287 | 0 | 0 | 8.931774 | 0 | 0 | 0 | 0 | 0 |
| 6 | 0 | 0 | 5.405394 | 0 | 0 | 2.951092 | 0 | 3.797284 | 4.576391 | 0 | 0 | 0 | 0 | 0 | 2.763756 | 0 | 0 | 2.562996 |
| 7 | 0 | 0 | 0 | 8.068472 | 0 | 3.267916 | 0 | 0 | 5.09157 | 6.082168 | 0 | 0 | 5.42044 | 0 | 0 | 4.431955 | 0.415844 | 2.73227 |
| 8 | 6.169271 | 4.918356 | 5.566813 | 0 | 0 | 4.884737 | 5.168666 | 0 | 5.189289 | 0 | 0 | 0 | 2.49011 | 0 | 4.750784 | 2.994664 | 0 | 0 |
| 9 | 11.64548 | 0 | 0 | 12.16663 | 0 | 8.407408 | 0 | 0 | 0 | 0 | 0 | 0 | 4.289772 | 0 | 0 | 4.652006 | 0 | 0 |
| 10 | 9.957874 | 7.829115 | 0 | 0 | 0 | 0 | 0 | 0 | 7.082694 | 8.388349 | 0 | 0 | 0 | 0 | 0 | 4.706276 | 0 | 0.705345 |
| 11 | 9.994487 | 12.3192 | 3.058695 | 0 | 0 | 0 | 6.111047 | 0.380701 | 3.904454 | 0 | 2.573056 | 0 | 0 | 0 | 0 | 5.610187 | 0 | 0 |
| 12 | 0 | 8.446147 | 7.506574 | 0 | 0 | 5.846259 | 7.362241 | 6.557457 | 7.627757 | 9.05184 | 0 | 0 | 0 | 0 | 6.646436 | 0 | 0 | 0 |
| 13 | 13.65315 | 18.11681 | 2.457055 | 0 | 8.218276 | 0 | 5.689919 | 0 | 4.45029 | 3.213032 | 5.992753 | 0 | 11.56691 | 0 | 0 | 7.734966 | 0 | 0 |
| 14 | 0 | 0 | 0 | 8.710629 | 0 | 0 | 0 | 0 | 6.466624 | 0 | 0 | 0 | 3.865449 | 0 | 5.339944 | 3.943355 | 0 | 0 |
| 15 | 11.08665 | 0 | 0 | 12.57808 | 0 | 8.377558 | 0 | 9.269582 | 0 | 10.28637 | 0 | 0 | 4.141793 | 0 | 0 | 4.953615 | 0 | 0.433766 |
| 16 | 0 | 0 | 7.32989 | 9.848915 | 0 | 0 | 6.639803 | 0 | 0 | 0 | 0 | 0 | 0 | 0 | 0 | 4.288343 | 0 | 0 |
| 17 | 0 | 0 | 8.49376 | 0 | 0 | 0 | 7.403671 | 9.346368 | 0 | 0 | 0 | 0 | 0 | 0 | 0 | 0 | 0 | 0 |
| 18 | 9.52255 | 0 | 0 | 10.30969 | 0 | 0 | 6.508697 | 0 | 0 | 9.04743 | 0 | 0 | 3.113288 | 0 | 7.667032 | 0 | 0 | 0 |
| 19 | 0 | 9.237608 | 3.488988 | 7.443493 | 0 | 0 | 0 | 0 | 0 | 0 | 0.921821 | 1.305681 | 0 | 0 | 0 | 4.447716 | 0 | 4.174564 |
| 20 | 0 | 16.78071 | 2.745921 | 0 | 5.606468 | 0 | 7.824948 | 0 | 0 | 0 | 4.807075 | 4.454489 | 0 | 0 | 0 | 7.226364 | 0 | 10.62363 |
| 21 | 0 | 0 | 8.18506 | 0 | 0.469365 | 4.241147 | 0 | 5.823779 | 0 | 0 | 0 | 0 | 0 | 0 | 0 | 6.475445 | 0 | 4.49432 |
| 22 | 9.609696 | 12.07202 | 0 | 6.483721 | 0 | 0 | 0 | 0 | 0 | 0 | 1.554688 | 0 | 5.446015 | 0 | 0 | 0 | 0 | 9.85667 |
| 23 | 10.71318 | 0 | 0 | 11.44685 | 0 | 8.097867 | 0 | 8.832153 | 8.646919 | 0 | 0 | 0 | 0 | 0 | 0 | 4.705225 | 0 | 0 |
| 24 | 6.625456 | 0 | 3.686915 | 6.715843 | 0.187058 | 0 | 3.735899 | 3.55698 | 0 | 0 | 0 | 0 | 0 | 0 | 2.996381 | 3.700704 | 0 | 0 |
| 25 | 9.794333 | 0 | 0 | 9.788224 | 0 | 4.599581 | 0 | 0 | 0 | 0 | 0 | 0 | 0 | 0 | 0 | 4.694789 | 0 | 0 10 |
| 26 | 10.25995 | 0 | 0 | 9.531824 | 0 | 1.156152 | 6.604298 | 0 | 0 | 0 | 0 | 0 | 6.346496 | 0 | 1.300262 | 0 | 1.869395 | 4.265034 |

Fig. 6. Multidimensional data with difficulty for visual pattern discovery.

However, a common ML modeling practice (without visualizing the data) starts with a simplest model, which is a linear discrimination function (black line in Fig. 5) to separate the blue and red points. It will fail. In contrast, visualization immediately gives an insight of a correct model class of "crossing" two linear functions, with one line going over blue points, and another one going over the red points.

How to reproduce such success in 2-D for n-D data such as shown in Fig. 6 where we cannot see a visual pattern in the data with a naked eye. The next section presents methods for lossless and interpretable visualization of n-D data in 2-D.

## 4. Visual discovery of ML models

### 4.1. Lossy and lossless approaches to Visual Discovery in n-D data

Visual discovery in n-D data needs to represent n-D data visually in the form, which will allow to discover **undistorted** n-D patterns in 2-D. Unfortunately, in high dimensions one cannot comprehensively see data. Lossless and interpretable visualization of n-D data in 2-D is required to *preserve multidimensional properties* for discovering undistorted ML models and their explanation.

Often multidimensional data are visualized by **lossy dimension reduction** (e.g., Principal Component Analysis), where each n-D point is mapped to a single 2-D point, or by **splitting** n-D data into a set of low dimensional data (pairwise correlation plots). While splitting is useful it *destroys integrity* of n-D data, and leads to a shallow understanding complex n-D data.

An alternative, for deeper understanding of n-D data is visual representations of n-D data in low dimensions **without splitting** and **loss of information**, is **graphs not 2-D points**, e.g., Parallel and Radial coordinates.

Fig. 7 illustrates the difference in the approaches.

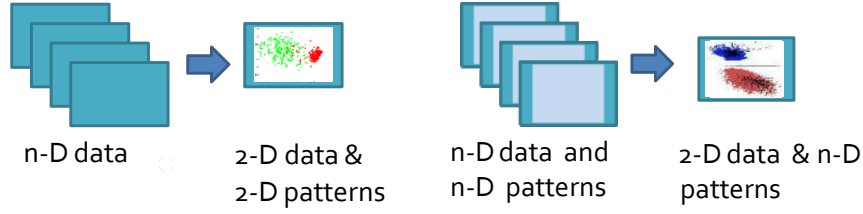

(a) Conversion of n-D data to 2-D with loss of some n-D information and visual discovery of distorted n-D patterns in 2-D data.

(b) n-D data are converted to 2-D without loss of information and abilities for visual discovery of undistorted n-D patterns in 2-D.

Fig. 7. The difference between lossy and lossless approaches.

### 4.2. Theoretical Limitations

The source of information loss in the process of dimension reduction from $n$ dimensions to $k$ dimensions ($k<n$) is in the smaller neighborhoods in k-D when each n-D point is mapped to k-D point. In particular, the 2-D/3-D visualization space (with $k=2$ or $k=3$) does not have enough neighbors to represent the n-D distances in 2-D. For instance, the 3-D binary cube has $2^3$ nodes, but the 10-D hypercube has $2^{10}$ nodes. Mapping $2^{10}$ 10-D points to $2^3$ 3-D points leads to the distortion of n-D distances, because the variability of distances between 3-D points is much smaller than between 10-D points. It leads to the significant **corruption** of n-D distances in 2-D visualization. The Johnson-Lindenstrauss lemma states these differences explicitly. It implies that only *a small number* of **arbitrary** n-D points can be mapped to k-D points of a smaller dimension $k$ that *preserves n-D distances with relatively small deviations.*

**Johnson-Lindenstrauss Lemma** [17].
Given $0 < \varepsilon < 1$, a set $X$ of $m$ points in $R^n$, and a number $k > 8\ln(m)/\varepsilon^2$, there is a linear map $f : R^n \to R^k$ such that for all $u, v \in X$.
$$(1-\varepsilon)\|u-v\|^2 \leq \|f(u)-f(v)\|^2 \leq (1+\varepsilon)\|u-v\|^2.$$

In other words, this lemma sets up a relation between $n$, $k$ and $m$ when the distance can be preserved with some allowable error $\varepsilon$. A version of the lemma [6] defines the possible dimension$s$ $k < n$, such that for any set of $m$ points in $R^n$ there is a mapping $f: R^n \to R^k$ with "similar" distances in $R^n$ and $R^k$ between mapped points. This similarity is expressed in terms of the error $0 < \varepsilon < 1$.

For $\varepsilon=1$, the distances in $R^k$ are less or equal to $\sqrt{2}\,S$, where $S$ is the distance in $R^n$. This means that the distance $s$ in $R^k$ will be in the interval $[0, 1.42S]$. In other words, the distances will not be more than 142% of the original distance, i.e., it will not be much exaggerated. However, it can dramatically diminish to 0. The lemma and this theorem allow to derive three formulas, to estimate the number of dimensions (sufficient and insufficient) to support the given distance errors. These formulas show that to keep distance errors within about 30%, for just 10 arbitrary high-dimensional points, the number of dimensions $k$ needs be over 1900 dimensions, and over 4500 dimensions for 300 arbitrary points. The point-to-point visualization methods do not meet these requirements for arbitrary datasets. Thus, this lemma sets up the theoretical limits to preserve n-D distances in 2-D. For details, see [Kovalerchuk, 2020].

### 4.3. Examples of Lossy vs. Lossless approaches for Visual Model Discovery

### 4.3.1. GLC-L Algorithms for Lossless Visual Model Discovery

The GLC-L algorithm [Kovalerchuk, Dovhalets, 2017] allows lossless visualization of n-D data and discovering a classification model. It is illustrated first for a lossless visualization of 4-D point **x**=( $x_1$, $x_2$, $x_3$, $x_4$ )=(1, 0.8, 1.2, 1) in Fig 8. The **algorithm** for this figure consists of the following steps:
- Set up 4 coordinate lines at different angles $Q_1$-$Q_4$
- Locate values $x_1$-$x_4$ of 4-D point **x** as blue lines (vectors) on respective coordinate lines
- Shifting and stacking blue lines
- Projecting the last point to U line
- Do the same for other 4-D points of blue class
- Do the same for 4-D points of red class
- Optimize angles $Q_1$-$Q_4$ to separate classes (yellow line).

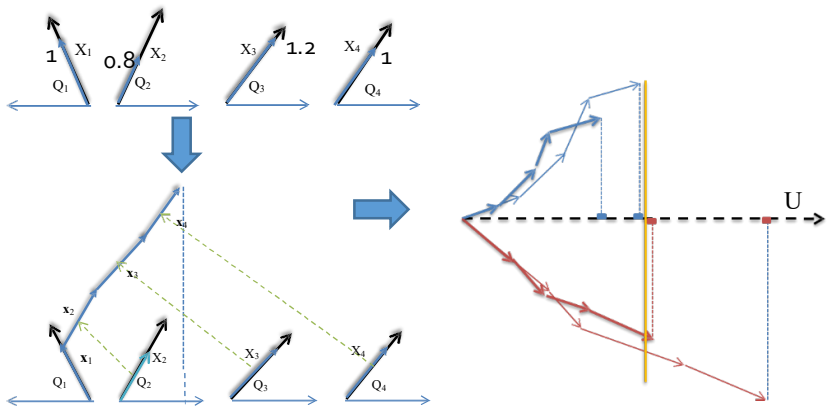

Fig. 8. GLC-L Algorithms for lossless Visual Model discovery

The applicability of the GLC-L algorithm to important tasks is illustrated in Fig. 9 for 9-D breast cancer diagnostics task using Wisconsin Breast Cancer data from the UCI ML repository. It allowed explanation of patterns and visual understanding of them, with lossless reversible/restorable visualization. It reached high accuracy with only one malignant (red case) on the wrong side.

The resulting linear classification model can be converted to a set of interpretable logical rules as was described in Section 1.2 by building a respective step function.

Also, this example shows a *fundamentally new opportunity* for splitting data into training and validation sets. This is a critical step of any ML model justification -- checking accuracy of the model. Traditionally in ML, we *split data randomly* to training and validation sets, compute accuracy on each of them, and if they are similar and large enough for the given task, we accept the model as a predictive tool.

While this is a common approach, which is used for decades, it is not free from several deficiencies [Kovalerchuk, 2020b]. First, we cannot explore all possible splits of a given set of n-D data points, because the number of splits is growing exponentially with the size of the dataset. As a result, we use a randomly selected small fraction of all splits, such as 10-fold Cross-Validation (CV). This is not only a small fraction of all splits, the selected splits overlap as in 10-fold CV, respectively, the accuracy estimates are not independent. As a result, we can get a biased estimate of the model accuracy, and make a wrong decision to accept or reject the model as a predictive tool.

Why do we use such random splits, despite these deficiencies? The reason is that we *cannot see multidimensional data with the naked eye,* and we are forced to use a random split of data into training and validation datasets. This is clear from the example in Fig. 3 above, which represents 2-D data. If all training cases will be selected, as cases below the red line, and all validation cases will be cases above that line, we will get a biased accuracy. Visually we can immediately see this bias in 2-D, for 2-D data. In contrast we cannot see it in n-D space, for n-D data.

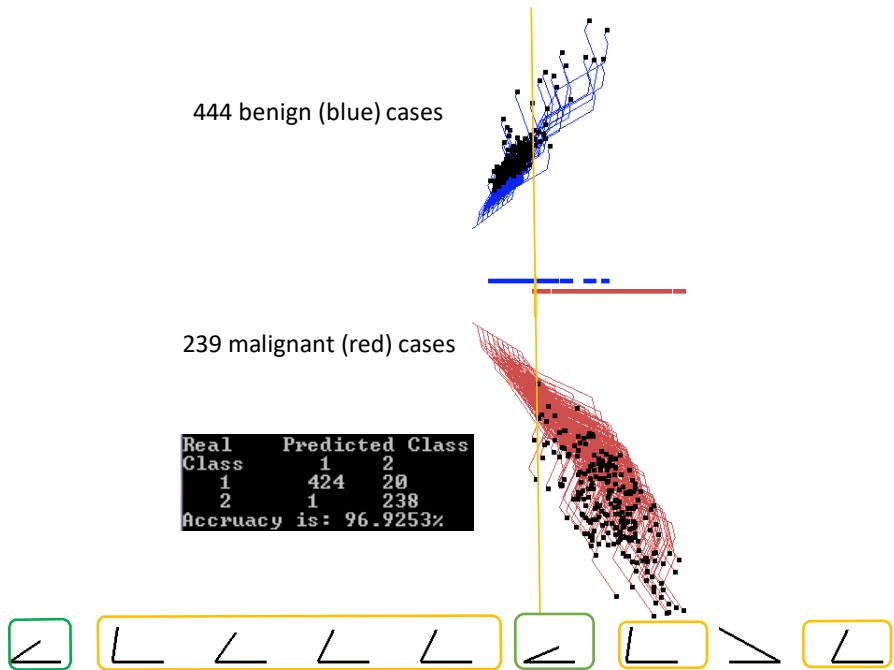

Fig. 9. WBC classification model. Angles in green boxes are most informative.

### 4.3.2. Avoiding Occlusion with Deep Learning

This example uses the same WBC data as above. In Fig. 9 polylines that represent different cases occlude each other, making discovering visual patterns by a naked eye challenging. In [Kovalerchuk, Dovhalets, 2017] the occlusion is avoided, by using a combination of GLC-L algorithm, described above, and a Convolutional Neural Network (CNN) algorithm. The first step is converting non-image WBC data to images, by GLC-L, and the second one is discovering a classification model, on these images by CNN. Each image represents a single WBC data case, as a single polyline (graph) com-pletely avoiding the occlusion. n. It resulted in 97.22% accuracy on 10-fold cross validation [Kovalerchuk, Dovhalets, 2017]. If images of n-D points are not compressed, then this combination of GLC-L and CNN is lossless.

## 5. General Line Coordinates (GLC)

### 5.1. General Line Coordinates to convert n-D points to graphs

General Line Coordinates (GLC) [Kovalerchuk, 2018] break a 400-year-old tradition of using the orthogonal Cartesian coordinates, which fit well to modeling the 3-D physical world, but are limited, for lossless visual representation of the diverse and abstract high-dimensional data, which we deal with in ML. GLC *relax the requirement of orthogonality*.

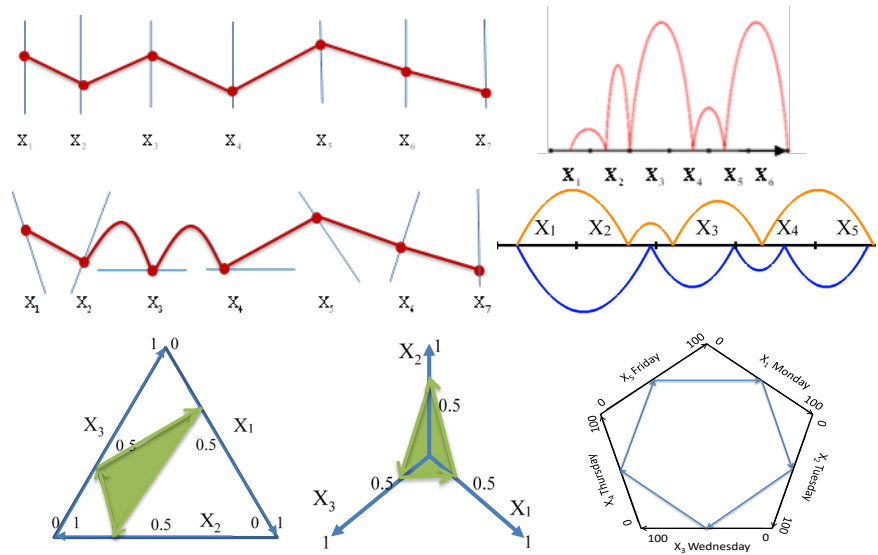

Fig. 10. Examples of different GLCs: Parallel, Non-parallel, Curved, In-line Coordinates, Triangular, Radial and Pentagon Coordinates.

In GLC, the points on the coordinates form **graphs**, where coordinates can overlap, collocate, be connected or disconnected, straight or curvy and go into any direction. Figs. 10-12 show the examples of several 2-D GLC types and Fig. 13 shows different ways how GLC graphs can be formed. The case studies in the next section show benefits of GLC for ML. Table 1 outlines 3-D GLC types. Several GLCs are described in more details in the next section in case studies. For full description of GLCs see [Kovalerchuk, 2018]. GLCs are **interpretable** because they use original attributes from the domain and do not construct artificial attributes that are foreign to the domain and it is done in methods such as PCA and t-SNE.

**Table 1. General Line Coordinates (GLC): 3-D visualization.**

| Type | Characteristics |
|---|---|
| 3-D General Line Co-ordinates (GLC) | Drawing $n$ coordinate axes in 3-D in variety of ways: curved, parallel, unparalleled, collocated, disconnected, etc. |
| Collocated Tripled Coordinates (CTC) | Splitting $n$ coordinates into triples and representing each triple as 3-D point in the same three axes; and linking these points to form a directed graph. If n mod 3 is not 0 then repeat the last coordinate $X_n$ one or two times to make it 0. |
| Basic Shifted Tripled Coordinates (STC) | Drawing each next triple in the shifted coordinate system by adding (1,1,1) to the second tripple, (2,2,2) to the third tripple ($i$-1, $i$-1,$i$-1) to the $i$-th triple, and so on. More generally, shifts can be a function of some parameters. |
| Anchored Tripled Co-ordinates (ATC) in 3-D | Drawing each next triple in the shifted coordinate system, i.e., coordinates shifted to the location of the given triple of (anchor), e.g., the first triple of a given n-D point. Triple are shown relative to the anchor easing the comparison with it. |
| 3-D Partially Collocated Coordinates (PCC) | Drawing some coordinate axes in 3-D collocated and some coordinates not collocated. |
| 3-D In-Line Coordinates (ILC) | Drawing all coordinate axes in 3D located one after another on a single straight line. |
| In-Plane Coordinates (IPC) | Drawing all coordinate axes in 3D located on a single plane (2-D GLC embedded to 3-D). |
| Spherical and polyhedron coordinates | Drawing all coordinate axes in 3D located on a sphere or a polyhedron. |
| Ellipsoidal coordinates | Drawing all coordinate axes in 3D located on ellipsoids. |
| GLC for linear functions (GLC-L) | Drawing all coordinates in 3D dynamically depending on coefficients of the linear function and value of n attributes. |
| Paired Crown Coordinates (PWC) | Drawing odd coordinates collocated on the closed convex hull in 3-D and even coordinates orthogonal to them as a function of the odd coordinate value. |

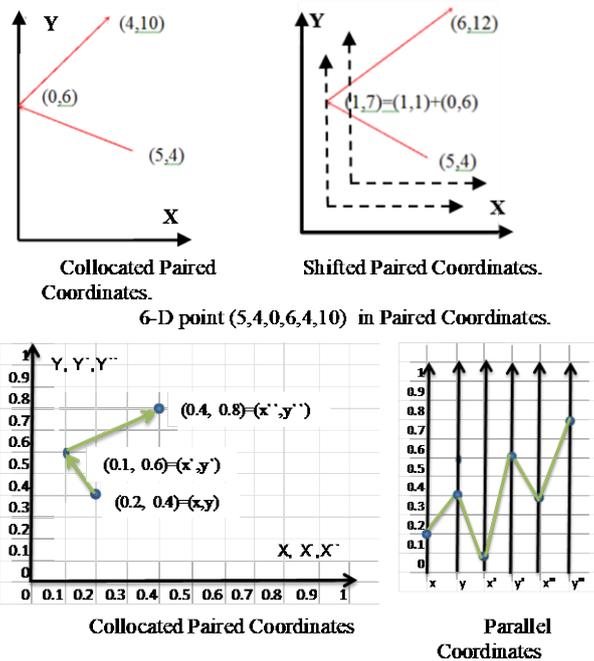

Fig. 11. Examples of the different GLCs: Collocated, Shifted Paired Coordinates, and Parallel Coordinates.

**Traditional Radial Coordinates (Radial Stars)** locate *n* coordinates radially and put *n* nodes on respective coordinates X*i* to represent each n-D point. Then these points are connected to form a "star".

The **Paired Radial Coordinates (CPC-Stars)** use a half of the nodes to get a reversible/lossless representation of an n-D point. It is done by creating n/2 radial coordinate axes and collocating coordinate $X_2$ with $X_3$, $X_4$ with $X_5$, and finally $X_n$ with $X_1$. Each pair of values of coordinates $(x_j, x_{j+1})$ of an n-D point **x** is displayed in its own pair of coordinates $(X_j, X_{j+1})$ as a 2-D point, then these points are connected to form a directed graph.

Fig. 12 shows data in these coordinates on the first row for 6-D data, and for 192-D. These coordinates have important advantages over traditional star coordinates for shape perception. Fig. 12 illustrates it showing the same 192-D data in both.

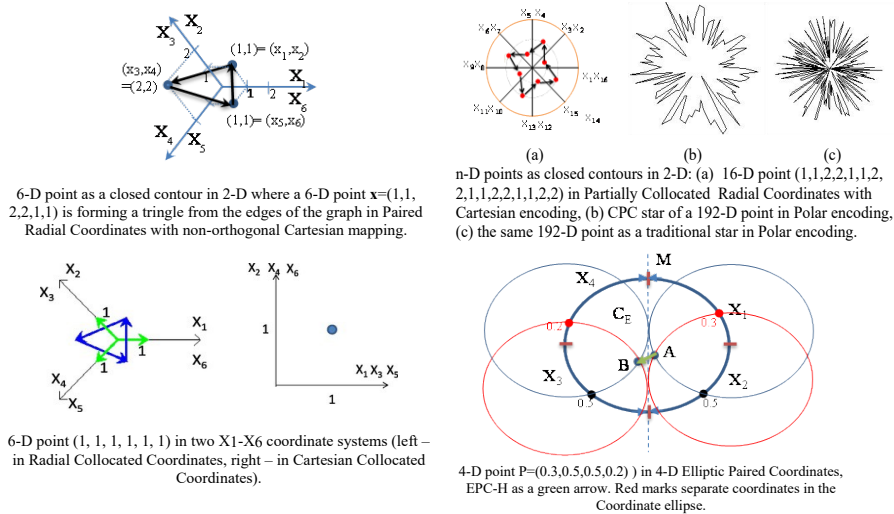

6-D point as a closed contour in 2-D where a 6-D point **x**=(1,1, 2,2,1,1) is forming a tringle from the edges of the graph in Paired Radial Coordinates with non-orthogonal Cartesian mapping.

n-D points as closed contours in 2-D: (a) 16-D point (1,1,2,2,1,1,2, 2,1,1,2,2,1,1,2,2) in Partially Collocated Radial Coordinates with Cartesian encoding, (b) CPC star of a 192-D point in Polar encoding, (c) the same 192-D point as a traditional star in Polar encoding.

6-D point (1, 1, 1, 1, 1, 1) in two $X_1$-$X_6$ coordinate systems (left – in Radial Collocated Coordinates, right – in Cartesian Collocated Coordinates).

4-D point P=(0.3,0.5,0.5,0.2) ) in 4-D Elliptic Paired Coordinates, EPC-H as a green arrow. Red marks separate coordinates in the Coordinate ellipse.

Fig. 12. Examples of different GLCs: Radial, Paired Collocated Radial, Cartesian Collocated and Elliptic Paired Coordinates.

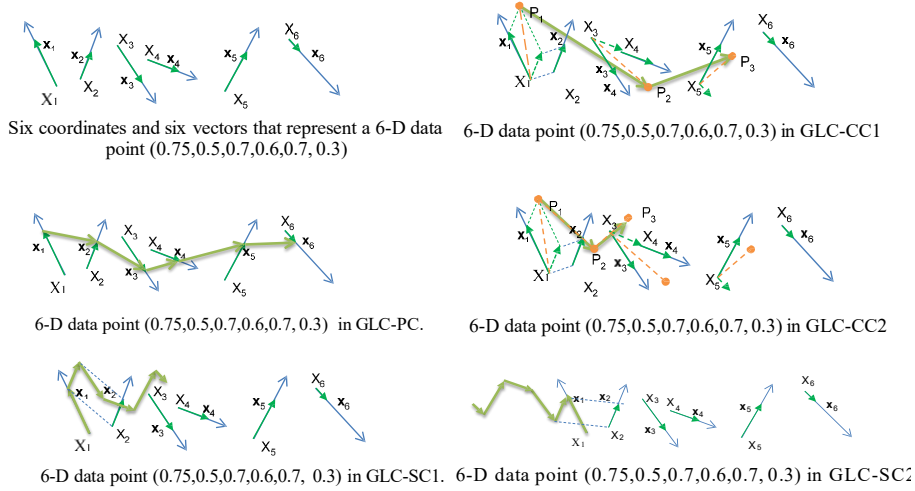

Fig. 13. Different ways to construct graphs of General Line Coordinates.

Several mathematical statements have been established for GLC [Kovalerchuk, 2018] that cover different aspects of the GLC theory and pattern simplification methodology. These statements are listed below. Fig. 14 illustrates some of these statements for the Shifted Paired Coordinates (SPC).

Statement 1. Parallel Coordinates, CPC and SPC preserve $L^p$ distances for $p=1$ and $p=2$, $D(x,y) = D^*(x^*,y^*)$.

Statement 2 (*n* points lossless representation). If all coordinates $X_i$ do not overlap, then GLC-PC algorithm provides bijective 1:1 mapping of any n-D point x to the 2-D directed graph x*.

Statement 3 (*n* points lossless representation). If all coordinates $X_i$ do not overlap then GLC-PC and GLC-SC1 algorithms provide bijective 1:1 mapping of any n-D point x to 2-D directed graph x*.

Statement 4 (*n/2* points lossless representation). If coordinates $X_i$, and $X_{i+1}$ are not collinear in each pair $(X_i, X_{i+1})$, then the GLC-CC1 algorithm provides bijective 1:1 mapping of any n-D point x to the 2-D directed graph x* with $\lceil n/2 \rceil$ nodes and $\lceil n/2 \rceil$ - 1 edges.

Statement 5 (*n/2* points lossless representation). If coordinates $X_i$, and $X_{i+1}$ are not collinear in each pair $(X_i, X_{i+1})$ then GLC-CC2 algorithm provides bijective 1:1 mapping of any n-D point x to 2-D directed graph x* with $\lceil n/2 \rceil$ nodes and $\lceil n/2 \rceil$ - 1 edges.

Statement 6 (*n* points lossless representation). If all coordinates $X_i$ do not overlap then GLC-SC2 algorithm provides bijective 1:1 mapping of any n-D point x to 2-D directed graph x*.

Statement 7. GLC-CC1 preserves $L^p$ distances for $p=1$,

$$D(x,y) = D^*(x^*,y^*).$$

Statement 8. In the coordinate system $X_1, X_2, \ldots, X_n$ constructed by the Single Point algorithm with the given base n-D point x=$(x_1, x_2, \ldots, x_n)$ and the anchor 2-D point *A*, the n-D point x is mapped one-to-one to a single 2-D point *A* by GLC-CC algorithm.

Statement 9 (locality statement). All graphs that represent nodes *N* of n-D hypercube *H* are within the square S.

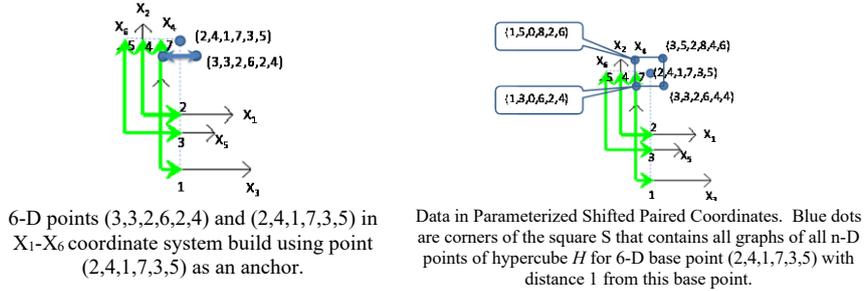

6-D points (3,3,2,6,2,4) and (2,4,1,7,3,5) in $X_1$-$X_6$ coordinate system build using point (2,4,1,7,3,5) as an anchor.

Data in Parameterized Shifted Paired Coordinates. Blue dots are corners of the square S that contains all graphs of all n-D points of hypercube *H* for 6-D base point (2,4,1,7,3,5) with distance 1 from this base point.

Fig. 14. Lossless visual representation of 6-D hypercube in Shifted Paired Coordinates.

### 5.2. Case Studies

#### 5.2.1. World Hunger data

To represent n-D data, in Collocated Paired Coordinates (CPC), we split an n-D point x into pairs of its coordinates $(x_1, x_2), \ldots, (x_{n-1}, x_n)$; draw each pair as a 2-D point in the collocated axes; and link these points to form a directed graph. For odd n coordinate $x_n$ is repeated to make *n* even. Fig. 15 shows advantages of visualization of the Global Hunger Index (GHI) for several countries in CPC over traditional time series visualization [Kovalerchuk, 2014; 2018]. This CPC visualization is simpler, and without occlusion and overlap of the lines.

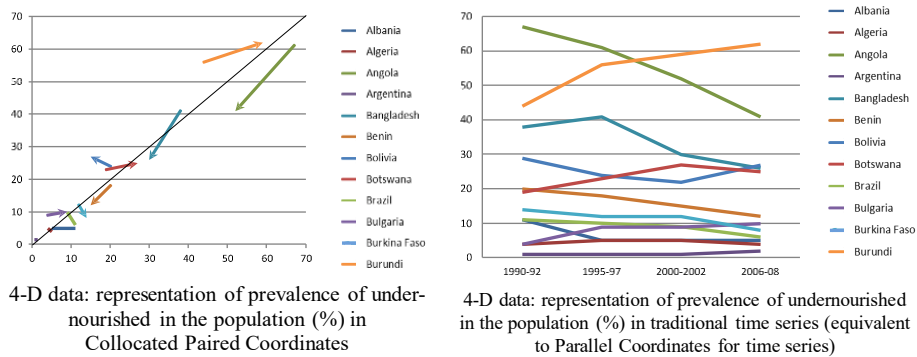

4-D data: representation of prevalence of undernourished in the population (%) in Collocated Paired Coordinates

4-D data: representation of prevalence of undernourished in the population (%) in traditional time series (equivalent to Parallel Coordinates for time series)

Fig. 15. Visualization of the Global Hunger Index (GHI) in Collocated Paired Coordinates (CPC) vs. traditional time series visualization.

### 5.2.2. Machine Learning for Investment Strategy with CPC

The goal of this study is learning trading investment strategy to predict long and short positions [Wilinski, Kovalerchuk, 2017]. It is done in 4-D and 6-D spaces, which are represented in Collocated Paired Coordinates (CPC), and Collocated Tripled Coordinates (CTC), respectively. In CPC, each 4-D point is an arrow in 2-D space (see previous section), and each 6-D point is an arrow in 3-D CTC space.

Each 2-D arrow consists of two pairs $(V_r, Y_r)$ of values (volume $V_r$ and relative main outcome variable $Y_r$) at two consecutive moments. In contrast with traditional timeseries CPC has no time axis. The arrow direction shows time from $i$ to $i+1$. The arrow beginning is the point in the space $(V_{r\,i}, Y_{r\,i})$, and its head is the next time point in the collocated space $(V_{r\,i+1}, Y_{r\,i+1})$.

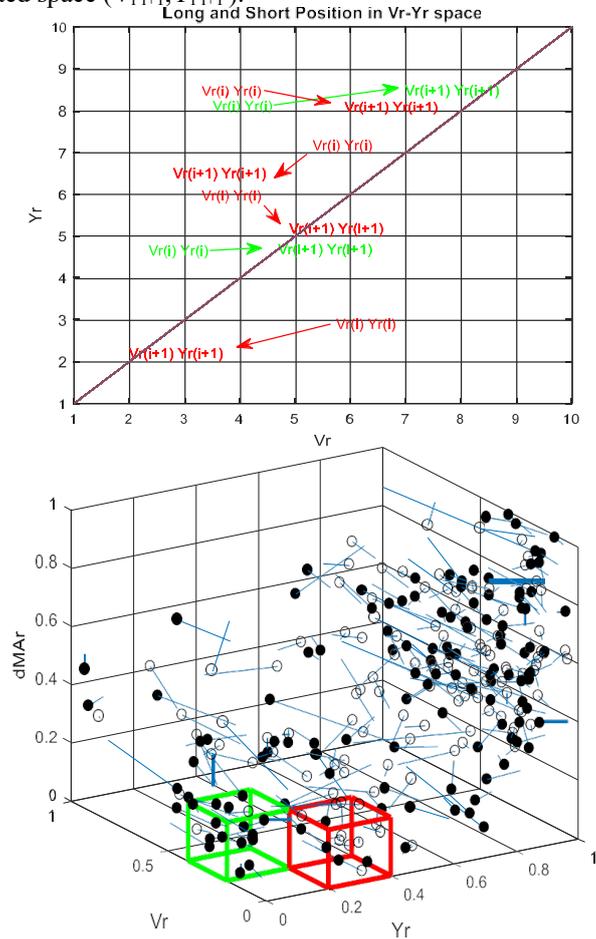

Fig.16. 4-D and 6-D trading data in 2-D and 3-D CPC with the maximum asymmetry between long (green) and short (red) positions [Wilinski, Kovalerchuk, 2017].

CPC give the inspiration idea for building a trading strategy, in contrast with the timeseries figure without it. It allows finding the areas with clusters of two kinds of arrows. In fig. 16, the arrows for the long positions are green arrows. The arrows for the short positions, are red. Along the $Y_r$ axis we can observe a type of change in Y in the current candle. if $Y_{r\,i+1} > Y_{r\,i}$ then $Y_{i+1} > Y_i$ the right decision in *i*-point is a long position opening. Otherwise, it is a short position.

Next, CPC shows the effectiveness of a decision in the positions. The very horizontal arrows indicate small profit. The more vertical arrows indicate the larger profit. In comparison with traditional time series, the CPC bring the additional knowledge about the potential of profit, in the selected area of parameters in $(V_r, Y_r)$ space.

The core of the learning process is searching squares and cubes in 2-D and 3-D CPC spaces with the prevailing number of long positions (green arrows). See Fig. 16. It is shown in [Wilinski, Kovalerchuk, 2017] that this leads to beneficial trading strategy in simulated training.

### 5.2.3. Recognition of digits with dimension reduction

Fig. 17 shows the results of GLC-L algorithm (see section 3.3) on MNIST handwritten digits [Kovalerchuk, Dovhalets, 2017]. Each image contains 22x22 = 484 pixels by cropping edges from original 784 pixels. The use of GLC-L algorithm allowed to go from 484-D to 249-D by the GLC-L algorithms, with minimal decrease of accuracy from 95.17% to 94.83%.

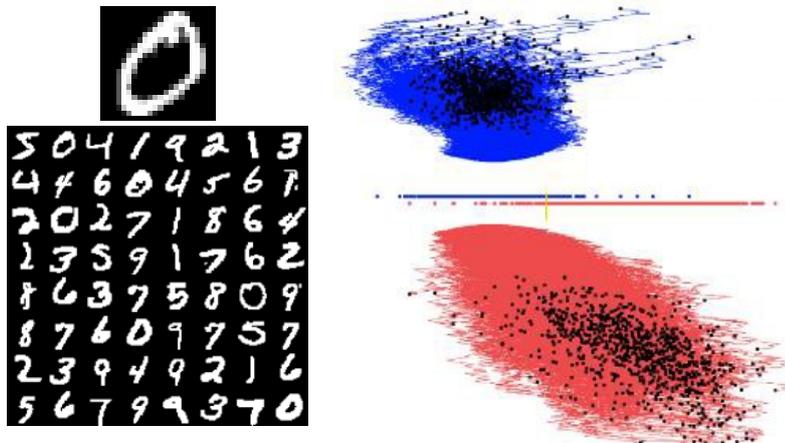

Fig.17. GLC-L algorithm for recognition of digits with dimension reduction.

### 5.2.4. Cancer Case Study with Shifted Paired Coordinates

This case study deals with the same 9-D WBC data, by using the FSP algorithm [Kovalerchuk B., Gharawi, 2018], and Shifted Paired Coordinates (SPC) [Kovalerchuk, 2018] for a graph representation of n-D points. The idea of SPC is presented in Fig. 18. The **Shifted Paired Coordinates (SPC)** visualization of the n-D data requires the splitting of *n* coordinates $X_1$-$X_n$ into pairs producing the *n*/2 non-overlapping pairs ($X_i,X_j$), such as $(X_1,X_2), (X_3,X_4), (X_5,X_6),…,(X_{n-1},X_n)$. In SPC, a pair $(X_i,X_j)$ is represented as a separate orthogonal Cartesian Coordinates (X,Y), where $X_i$ is X and $X_j$ is Y, respectively.

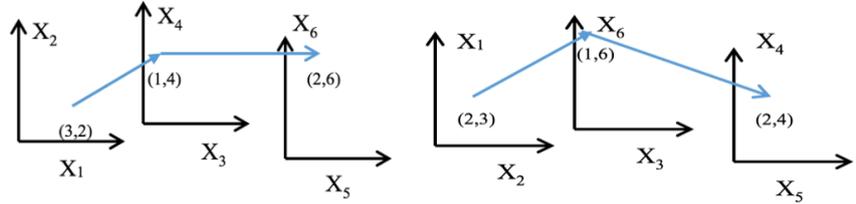

Point **a** in $(X_1,X_2)$, $(X_3,X_4)$, $(X_5,X_6)$ as a sequence of pairs (3,2), (1,4) and (2,6).

Point **a** in $(X_2,X_1)$, $(X_3,X_6)$, $(X_5,X_4)$ as a sequence of pairs (2,3), (1,6) and (2,4).

Fig. 18. 6-D point a=(3,2,1,4,2,6) in Shifted Paired Coordinates.

In SPC, each coordinate pair $(X_i,X_j)$ is *shifted* relative to other pairs to avoid their overlap. This creates $n/2$ scatter plots. Next, for each n-D point $x=(x_1,x_2,…,x_n)$, the point $(x_1,x_2)$ in $(X_1,X_2)$ is connected to the point $(x_3,x_4)$ in $(X_3,X_4)$ and so on until point $(x_{n-2},x_{n-1})$ in $(X_{n-2},X_{n-1})$ is connected to the point $(x_{n-1},x_n)$ in $(X_{n-1},X_n)$ to form a directed graph x*. Fig. 18 shows the same 6-D point, visualized in SPC, in two different ways, due to different pairing of coordinates.

The FSP algorithm has the three major steps: *Filtering* out the less efficient visualizations from the multiple SPC visualizations, *Searching* for sequences of paired coordinates that are more efficient for classification model discovery, and *Presenting* the model discovered with a best SPC sequence, to the analyst [Kovalerchuk, Gharawi, 2018]. The results of FSP applied to CPC graphs of WBC data are shows in Figs. 19-20.

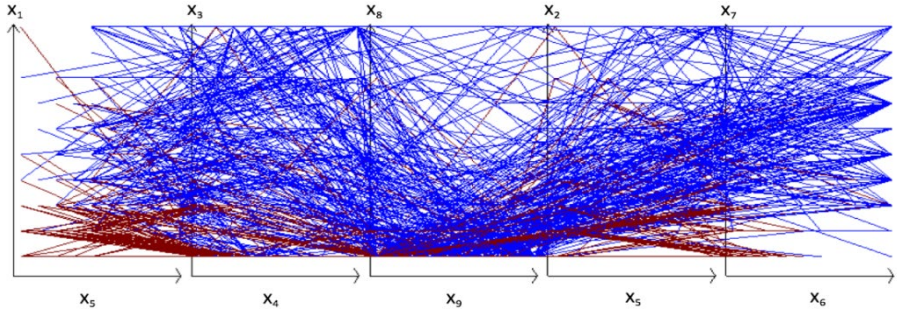

Fig. 19. Benign and malignant WBC data visualized in SPC as 2-D graphs of 10-D points.

Fig. 19 shows the motivation for filtering and searching in FSP. It presents WBC data in SPC, where graphs occlude each other making it difficult to discover the pattern visually. Fig. 20 shows the results of automatic filtering and searching by FSP algorithm. It displays only cases located outside of a small violet rectangle at the bottom in the middle and go inside of two larger rectangles on the left. These cases are dominantly cases of the blue class. Together these properties provide a rule:

If $(x_8,x_9) \in R_1$ & $(x_6,x_7) \notin R_2$ & $(x_6,x_7) \notin R_3$ then **x** $\in$ class Red else **x** $\in$ class Blue,

where $R_1$ and $R_2$ and $R_3$ are three rectangles described above. This rule has accuracy 93.60% on all WBC data [Kovalerchuk, Gharawi, 2018]. This fully interpretable rule is visual and intelligible by domain experts, because it uses only original domain features and relations.

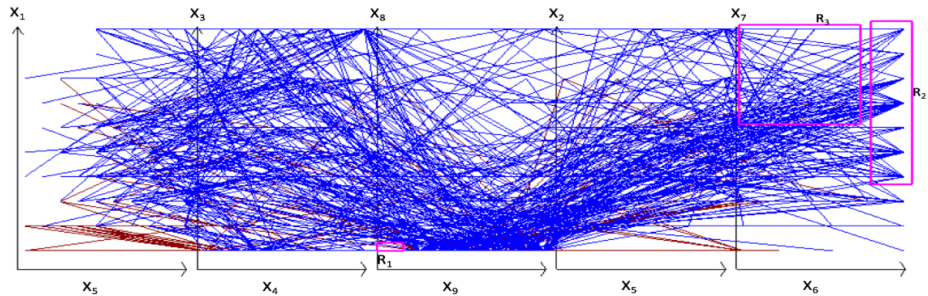

Fig. 20. SPC visualization of WBC data with areas dominated by the blue class.

This case study shows the benefits of combining analytical and visual means for producing interpretable ML models. The analytical FSP algorithm works on the multiple visual lossless SPC representations of n-D data, to find the interpretable patterns. While occlusion blocks discovering these properties by visual means, the analytical FSP algorithm discovers them in the SPC, simplifying the pattern discovery, providing the explainable visual rules, and decreasing the cognitive load.

### 5.2.5. Lossless visualization via CPC-Stars Radial Stars and Parallel Coordinates and human abilities to discover patterns in high-D data

The design of CPC-Stars vs. traditional Radial Stars was described in section 4.1. Several successful experiments have been conducted to evaluate human experts' abilities to discover visual n-D patterns [Grishin, Kovalerchuk, 2014; Kovalerchuk, Grishin, 2018, 2019; Kovalerchuk, 2018]. Fig. 21 shows lossless visualization of 48-D and 96-D data in CPC-Stars, Radial Stars and Parallel Coordinates.

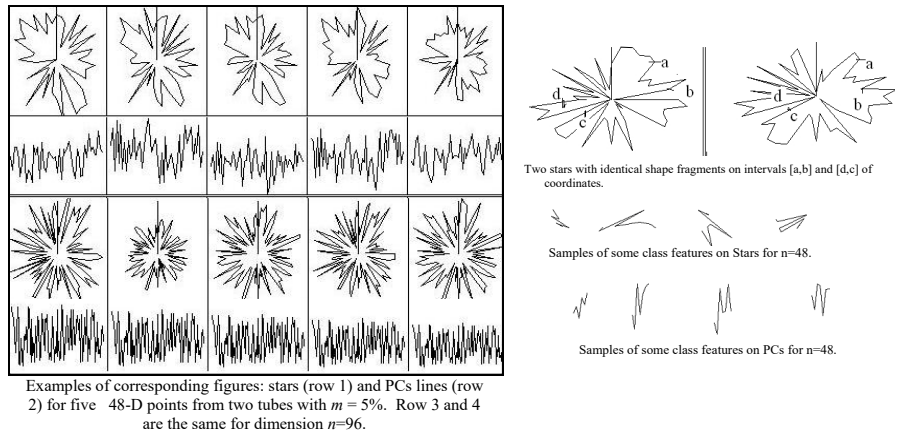

Examples of corresponding figures: stars (row 1) and PCs lines (row 2) for five 48-D points from two tubes with $m = 5\%$. Row 3 and 4 are the same for dimension $n=96$.

Two stars with identical shape fragments on intervals [a,b] and [d,c] of coordinates.

Samples of some class features on Stars for n=48.

Samples of some class features on PCs for n=48.

Fig. 21. 48-D and 96-D points in CPC-Stars, Radial Stars and Parallel Coordinates

While all of them are lossless, fully preserving high-dimensional data, abilities of humans to discover visual patterns shown on the right in Fig.21, are higher using CPC-stars, than using the two others. This is a result of using only $n/2$ nodes vs. $n$ nodes in alternative representations. Similar advantages have been demonstrated with 160-D, 170-D and 192-D. Fig. 22 shows the musk 170-D data from UCI ML repository.

The examples and case studies demonstrate that ML methods with General Line Coordinates allow: (1) **visualizing** data of multiple dimensions from 4-D to 484- D without loss of information and (2) **discovering interpretable** patterns by combining humans perceptual capabilities, and Machine Learning algorithms for classification of such high-dimensional data. Such hybrid technique can be developed further in multiple ways, to deal with different new challenging ML and data science tasks.

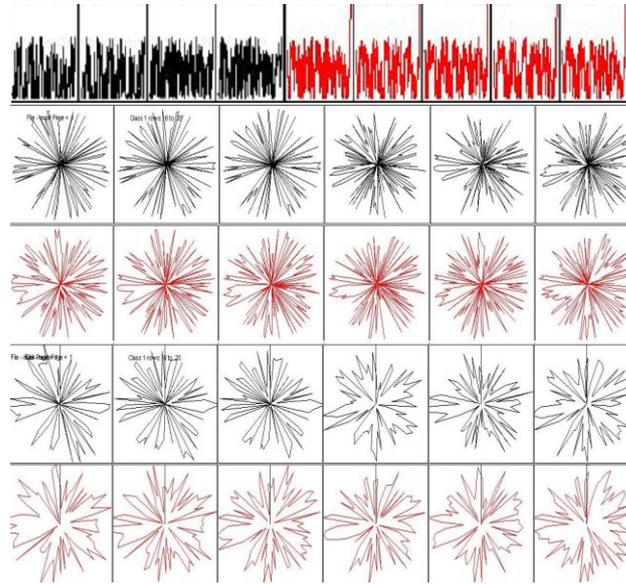

Fig. 22. Nine 170-dimensional points of two classes in Parallel Coordinates (row 1), in star coordinates (row 2 class "musk", row 3 class "non-musk chemicals"), and in CPC stars (row 4 class "musk" and row 5, class "non-musk chemicals").

## 6. Visual methods for Traditional Machine Learning

### 6.1. Visualizing association rules: matrix and parallel sets visualization for association rules

In [Zhang et al, 2019] association rules are visualized. A general form of the association rule (AR) is $A \Rightarrow B$, where $A$ and $B$ are statements. Commonly $A$ consists of several other statements, $A = P_1 \& P_2 \& \ldots \& P_k$, e.g.,

*If customers buy both tomato (T) and cucumbers (Cu), they likely buy carrots (Ca).*
Here $A = T\&Cu$ and $B = Cu$.

The qualities of the AR rule are measured by the *support* and *confidence* that express, respectively, *a frequency* of the itemset *A* in the dataset, and a portion of transactions with *A* and *B* relative to frequency of *A*. ARs are *interpretable* being a class of propositional rules expressed in the original domain terms.

The typical questions regarding ARs are as follows: What are the rules with the highest support/confidence? What are outliers of the rule and their cause? Why is the rule confidence low? What is the cause of rule? What rules are non-interesting? Visualization allows the answering of some of these questions directly. Fig. 23a shows structure-based visualization of association rules with a matrix and heatmap [Zhang et al, 2019]. Here left-hand sides (LHS) of the rules are on the right and right-hand sides (RHS) of the rules are on the top. Respectively, each row shows a possible LHS itemset of the rule, and each column shows a possible RHS itemset. The violet cells indicate discovered rules A⇒ B with respective LHS and RHS. The darker color of the rule cell shows a greater rule confidence. Similarly, the darker LHS shows the larger rule support. A similar visualization is implemented in the Sklearn package where each cell is associated with colored circles of different sizes to express the quality of the rule. The major challenges here are scalability and readability for a large number of LHS and RHS [Zhang et al, 2019]. A bird view solution was implemented in the Sklearn package, where each rule is a colored point in a 2-D plot with support and confidence as coordinates, which allowed to showing over 5000 rules.

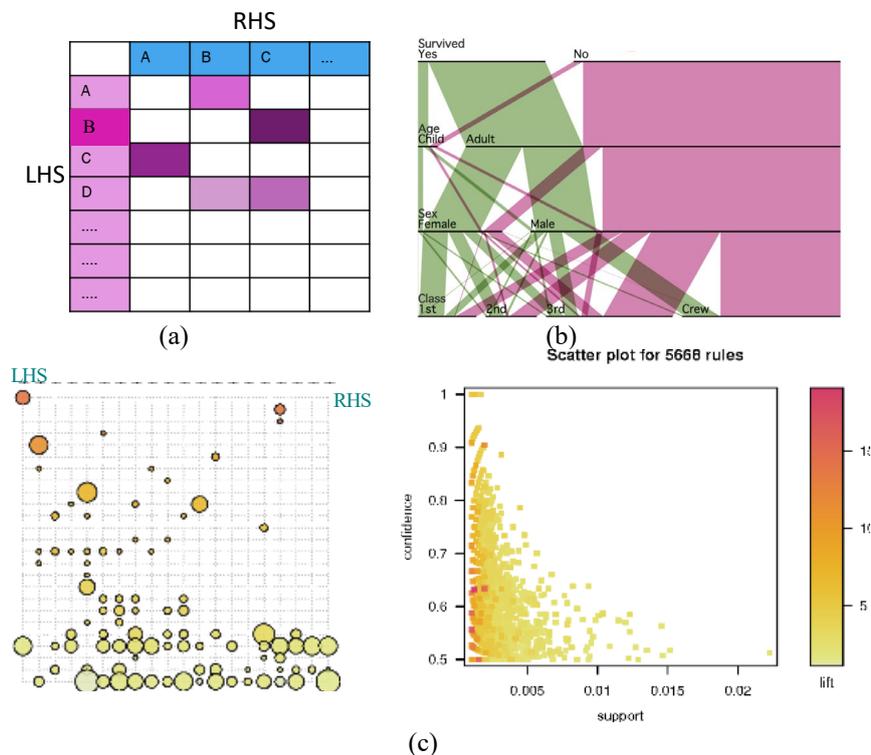

Fig. 23. Visualizations of association rules [Zhang et al, 2019] and Sklearn

Fig. 23b shows the input-based model visualization for a set of ARs. It uses Parallel Sets. Parallel Sets display dimensions as adjacent **parallel axes** and their values (categories) as **segments** over the axes. Connections between categories in the parallel axes form **ribbons.** The segments are like points, and the ribbons are like lines in Parallel Coordinates [Inselberg, 2009] The ribbon crossings cause clutter that can be minimized by reordering coordinates and other methods [Zhang et al, 2019]. Both model visualizations shown in Fig. 23 are valid for other rules-based ML models too.

### 6.2. Dataflow tracing in ML models: Decision Trees

Graph Visualizer, TensorBoard, TensorFlow's dashboard, Olah's interactive essays, ConvNetJS, TensorFlow Playground, and Keras are current tools for DNN dataflow visualization [Wongsuphasawat et al, 2018]. They allow observing scalar values, distribution of tensors, images, audio and others to optimize and understand models by visualizing model structure at different levels of detail.

While all these tools are very useful, the major issue is that dataflow visualization itself *does not explain or optimize* the DNN model. An experienced data scientist should guide data flow visualization for this. In contrast, the dataflow for explainable models can *bring explanation itself*, as we show below for Decision Trees (DTs).

Tracing the movement of a given n-D point in the DT shows all the interpretable decisions made to classify this point. For instance, consider a result of tracing the 4-D point **x**=(7,2,4,1) in the DT through a sequence of nodes for attributes $x_3, x_2, x_4, x_1$ with the following thresholds: $x_3 < 5$, $x_2 > 0$, $x_4 < 5$, $x_1 > 6$ to a terminal node of class 1. The point **x** satisfies all these directly interpretable inequalities.

(a) Traditional visualization of WBC data decision tree. Green edges and nodes indicate the benign class and red edges and nodes indicate the malignant class.

(b) DT with edges as Folded Coordinates in disproportional scales. The curved lines are cases that reach the DT malignant edge with different certainties due to the different distances from the threshold node.

Fig. 24. DT dataflow tracing visualizations for WBC data [Kovalerchuk, 2020].

Fig. 24 shows a traditional DT visualization for 9-D Wisconsin Breast Cancer (WBC) data from UCI Machine Learning repository. It clearly presents the structure of the DT model, but without explicitly tracing individual cases. The trace is added with a dotted polyline in this figure. Fig. 24b shows two 5-D points **a**= (2.8, 5, 2.5, 5.5, 6.5) and **b**= (5, 8, 3, 4, 6). Both points reach the terminal malignant edge of the DT, but with different certainty. The first point reaches it with a lower certainty, having its values closer to the thresholds of uc and bn coordinates.

In this visualization, called **Folded Coordinate Decision Tree (FC-DT)** visualization [Kovalerchuk, 2020], the edges of the DT not only connect decision nodes, but also serve as *Folded Coordinates* in disproportional scales for WBC data. Here, each coordinate is folded, at the node threshold point with different lengths of the sides. For instance, with threshold T=2.5 on the coordinate uc with the interval of values [1,10], the left interval is [1, 2.5), and the right interval is [2.5,10]. In Figure 29b, these two unequal intervals are visualized with equal lengths, i.e., forming a *disproportional scale*.

### 6.3. iForest: Interpreting Random Forests via Visual Analytics

The goal of iForest system [Zhao et l, 2018] is assisting a user in understanding *how* random forests make predictions and observing prediction quality. It is an attempt to open in the innerworkings of the random forests. To be a meaningful explanation for the end user it should not use terms, which are foreign for the domain where the data came from. Otherwise, it is an explanation for another user – the data scientist/ML model designer. The actual usability testing was conducted with this category of users (students and research scientists).

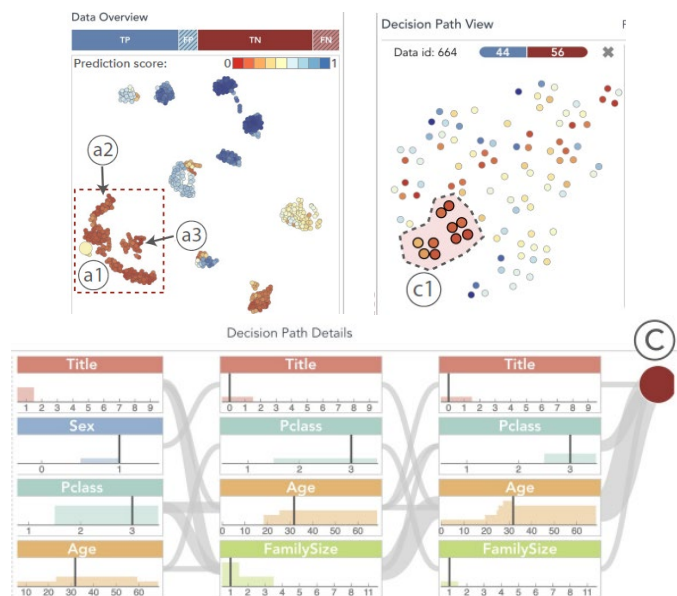

Fig 25. iForest to interpret random forests for Titanic data [Zhao et l, 2018].

iForest uses t-SNE to project data onto a 2D plane (Fig. 25) for data overview and analysis of similarity of the decision passes. We already discussed challenges of t-SNE in section 6 for such tasks. This challenge is illuminated in Fig 25 on the left where the yellow section a1 is identified as outlier from the classification viewpoint (low confidence to belong to the same class as t-SNE neighbors), but they are not outliers in t-SNE. Similarly, in Fig 30 on the right, multiple cases from different classes and of different confidence are t-SNE neighbors.

The reasons for this are that (1) t-SNE is an unsupervised clustering method that can differ from given classes, (2) t-SNE dense 2-D areas may not be dense areas in n-D [Maaten, 2018], and that t-SNE is a point-to-point mapping of n-D points to 2-D points with the loss of n-D information. This issue was discussed in prior sections in depth.

The major advantage of using t-SNE and the other point-to-point mappings of n-D data to 2-D data is that they suffer much less from *occlusion* than point-to- graph mapping which we discussed in the prior sections that are General Line Coordinates. In summary, the general framework of iForest is beneficial for ML model explanation and can be enhanced with point-to-graph methods that preserve n-D information in 2-D.

### 6.4. TreeExplainer for Tree Based Models

TreeExplainer is a framework to explain random forests, decision trees, and gradient boosted trees [Lundberg et al, 2019]. Its polynomial time algorithm computes explanations based on game theory and part of the SHAP (SHapley Additive exPlanations) framework. To produce an explanation, effects of local feature interaction are measured. Understanding global model structure is based on combining the local explanations of each prediction.

Fig. 26 illustrates its "white box" local explanation. As we see it deciphers mortality rate 4 as a sum 2.5+0.5+3-2 of 4 named features. In contrast, the black box model produces only the mortality rate 4 without telling how it was obtained. The question is can we call sum 2.5+0.5+3-2 a "white box" or it is rather a black-box explanation. The situation here is like in section 2.5 with quasi-explanations. If a user did not get any other information beyond the numbers 2.5, 0.5, 3 and -2 for 4 attributes, then it is a rather **black-box explanation** not a white box explanation. In other words, the user has got a black box model *prediction* of 4, and a black box *explanation* of 4 without answers for the questions like: why should numbers 2.5, 0.5, 3 and -2 be accepted, what is their meaning in the domain, and why summation of them makes sense in the domain.

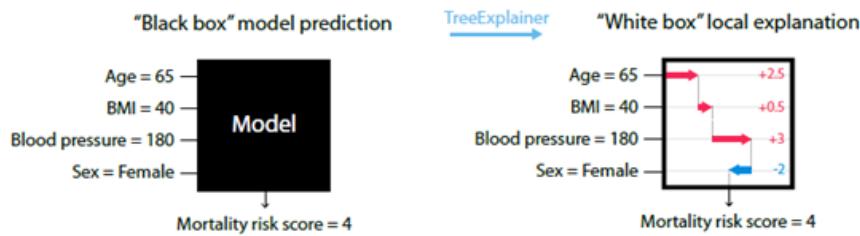

Fig. 26. TreeExplainer "white box" local explanation [Lundberg et al, 2019].

## 7. Traditional Visual Methods for Model Understanding: PCA, t-SNE and related point-to-point methods

Principal Component analysis (PCA), t-Distributed Stochastic Neighbor Embedding (t-SNE) [Maaten, 2018] and related methods are popular methods applied to get intuitive understanding of data, ML model and their relations.

These *point-to-point* projection methods convert n-D points to 2-D or 3-D points for visualization. PCA can *distort local neighborhoods* [Embeddings, 2019]. Often t-SNE attempts preserving local data neighborhoods, at the expense of *distorting global structure* [Embeddings, 2019]. Below we summarize and analyze the challenges and warnings with t-SNE highlighted in [Maaten, 2018; Embeddings, 2019]. including from t-SNE author. One of them is that t-SNE may not help to find *outliers* or assign meaning to point *densities* in clusters. Thus, **outliers** and **dense areas** visible in t-SNE may not be them in the original n-D space. Despite this warning, we can see the statements that users can easily identify outliers in t-SNE [Choo, Liu, 2018], and see the similarities [Zhao et l, 2018]. It will be valid only after showing that the n-D metrics are not distorted in 2-D for the *given data*. In general, for arbitrary *data*, any point-to-point dimension reduction method distorts n-D metrics in k-D of a lower dimension as shown in the Johnson-Lindenstrauss lemma presented above.

Fig. 27 shows PCA and t-SNE in 2-D, and 3-D visualizations of 81-D breast lesion ultrasound data [Jamieson et al, 2010]. These visualizations differ significantly creating very different opportunities to interpret data and models.

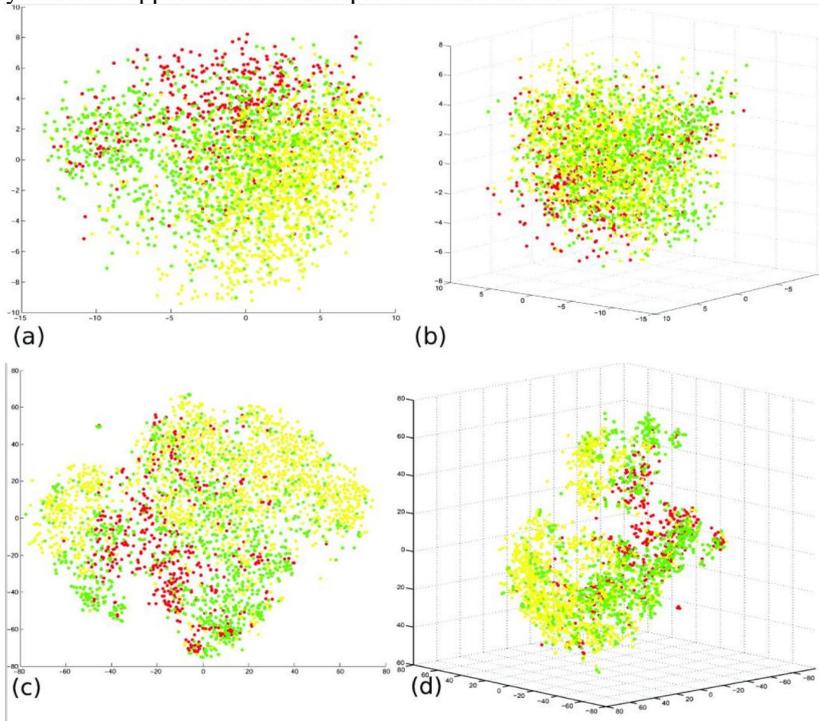

Fig. 27. PCA and t-SNE visualizations of 81-D breast lesion ultrasound data. Green is benign lesions, red is malignant, and yellow is benign-cystic. Here (a) 2-D and (b) 3-D PCA, and (c) 2-D and (d) 3-D t-SNE [Jamieson et al, 2010].

In this example, each 81-D point is compressed 40 times to get a 2-D point, and 27 times to get a 3-D point, respectively with significant **loss** of 81-D information. Each of these 4 visualizations captures different properties of the original 81-D data, and *losses* other properties. Moreover, properties presented in these visualizations are **artificial** "**summary**" properties, which differ from original interpretable attributes of 81-D points.

In fact, all PCA principal components have **no direct interpretation**, in the domain terms, for *heterogeneous attributes*. The same is true for t-SNE. Any attempts at discovering meaningful patterns in these 2/3-D visualizations will hit this wall of lack of direct interpretation of "summary" attributes. In general point-to-point methods like T-SNE and PCA, do not preserve all information of initial features (they are lossy visualizations of n-D data), and produce a "summary", which has no direct interpretation.

Another example with visualizations is explaining why the model misclassified some samples [Marino et al, 2018]. The idea is *generating a closest sample (from the correct class)* to the misclassified sample and visualizing the difference between their attributes as an *explanation of misclassification* (see Fig. 28c). This is a simple and an attractive way of explanation. Moreover, this new sample can be added to the training data to improve the model.

The goal of Fig. 28ab is *explaining visually the method* of finding the closest sample from the correct class proposed in [Marino et al, 2018]. It is done by visualizing these samples using t-SNE to see how close they are. Fig. 27a shows the original samples and Fig 27b shows them together with the closest samples from the correct class. It is visible that these samples are close (in fact they overlap) in t-SNE.

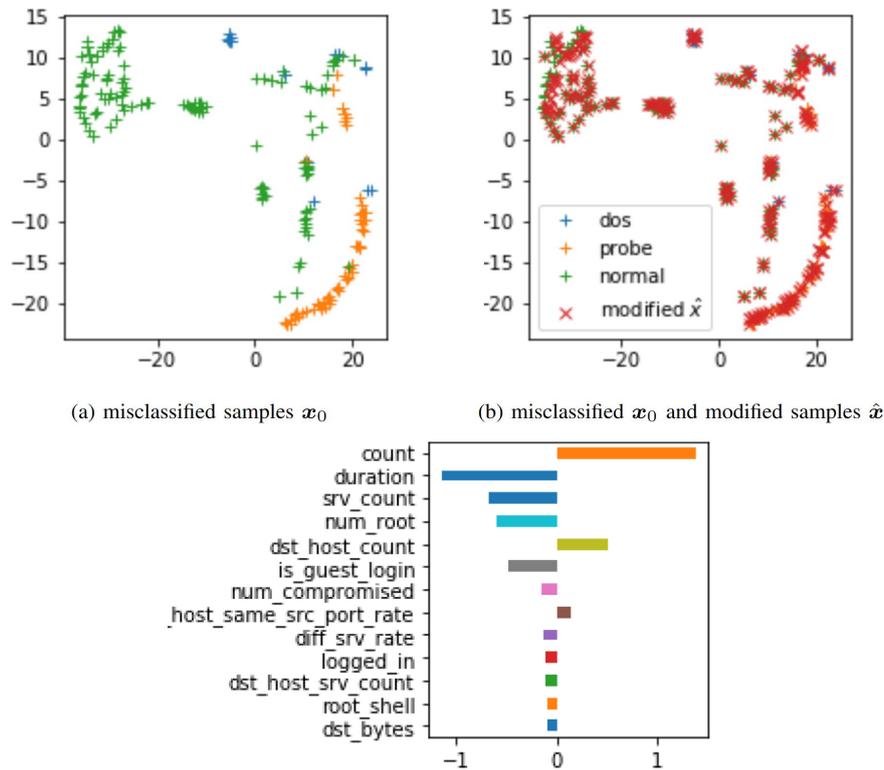

(a) misclassified samples $x_0$
(b) misclassified $x_0$ and modified samples $\hat{x}$

Normal samples misclassified as DOS using Linear model

Fig. 28. Misclassified and modified samples in t-SNE (a,b) with their differences (c) [Marino et al, 2018].

While the idea to explain the similarity of samples, by visualizing samples, has merit, the use of lossy point-to-point algorithms like t-SNE is not free from *deficiencies*, which can be resolved by applying lossless point-to-graph visualization methods such as GLC-L algorithm described in section 4. In Fig 28ab, t-SNE does not show the *difference between samples*, they overlap in t-SNE visualization. While it shows that the closest samples, computed by the algorithm from [Marino et al, 2018], and the t-SNE representation of these samples are quite consistent with each other, the resolution of t-SNE, for these data, is not sufficient to see the differences. Thus, t-SNE *distorted* the closeness, in the original n-D space computed in [Marino et al, 2018].

In addition, t-SNE does not show *alternative closest samples* from the correct classes. In n-D space, the number of closest samples is growing *exponentially* with the dimension growth. Which one to pick up for explanation? We may get attribute $x_i$=5 in one neighbor sample, and $x_i$=-5 in another one from the same correct class. Respectively, these neighbors will lead to opposite explanations. Averaging such neighbors to $x_i$=0 will nullify the contribution of $x_i$ to the explanation. In [Marino et al, 2018] a sample was selected, based on the proposed algorithm, without exploring alternatives.

The AtSNE algorithm [Fu et al, 2019] is to resolve the difficulties of t-SNE algorithm, for capturing the global n-D data structure, by generating 2-D anchor points (2-D skeleton) from the original n-D data with a hierarchical optimization. This algorithm is only applicable to the cases when the global structure of the n-D dataset can be captured by a planar structure using point-to-point mapping (n-D point to 2-D point). In fact, 2-D skeleton can corrupt the n-D structure (see the Johnson-Lindenstrauss lemma above). Moreover, the meaningful similarity between n-D points can be *non-metric*.

The fact, that t-SNE and AtSNE can distort n-D structures in a low-dimensional map, is the *most fundamental deficiency* of all point-to-point methods, along with the lack of interpretation of generated dimensions. Therefore, we focus on **point-to-graph** GLC methods, which open a new opportunity to address these challenges.

Fig 29 shows how GLC-L resolves the difficulties exposed in Fig. 28. It shows a 4-D misclassified case, and two nearest 4-D cases of the correct class, where dotted lines show the attributes, which changed values relative to the misclassified case. The vertical yellow line is a linear discrimination line, of the red and blue classes. This lossless visualization preserves all 4-D information, with interpretable original attributes. It does not use any artificial attributes.

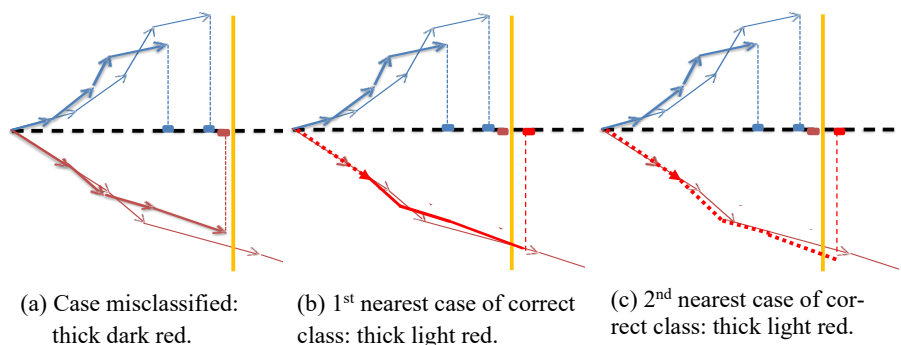

(a) Case misclassified: thick dark red.

(b) 1st nearest case of correct class: thick light red.

(c) 2nd nearest case of correct class: thick light red.

Fig. 29. GLC-L: misclassified case (a) and two nearest cases of correct class (b, c) with dotted lines show changed attributes.

**Interpreting Time Series**. Trends in retrospective time series data are relatively straightforward to understand. How do we understand predictions of time series in data? In [Schlegel et al, 2019] existing ML predictive algorithms and their explanation methods such as LRP, DeepLIFT, LIMR and SHAP are adapted for the specifics of timeseries. Time points $t_i$ are considered as features. Training and test data are sequences of *m* such features. Each feature is associated with its importance/relevance indicators $r_i$ computed by a respective explanation method. The vector of $r_i$ is considered as an explanation of the sequenc,e playing the same role as salience of pixels. Respectively they are also visualized by a heatmap (see Fig. 30. The authors modify test sequences in several ways (e.g., permuting sequences), and explore how the vector of explanations $r_i$ is changed. Together with domain knowledge, an expert can inspect the produced explanation visualizations. However, the abilities for this are quite limited in the same way as for salience of pixels that we discussed in section 2.5, because explanations $r_i$ are still black boxes for domain experts.

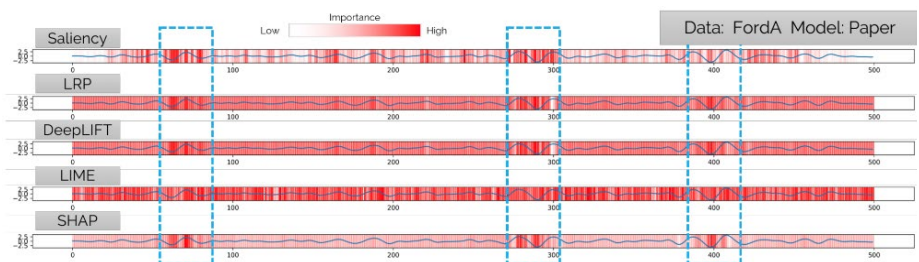

Fig. 30. Relevance heatmaps on an exemplary time series with different relevance/explanation indicators [Schlegel et al, 2019].

## 8. Interpreting Deep Learning

### 8.1. Understanding Deep Learning via Generalization Analysis

The most challenging property of DNN models is that the number of their parameters way *exceeds* the number of training cases. It makes overfitting and memorization of training data quite likely with the *failure to generalize* to the test data, outside of the training data.

There are empirical observations that DNN trained with stochastic gradient methods fit a random labeling of the training data even after replacing the true images by completely unstructured random noise. [Zhang et al, 2017]. It was expected by these authors that the learning should not be converging or slowing down, but for multiple standard architectures it did not happen. This is consistent with their theoretical result below. This theorem is for a finite sample of size *n*, and complements the NN universal approximation theorems that are for the entire domain.

<u>Theorem</u>. There exists a two-layer neural network with 2n+d weights that can represent any function on a sample of size *n* in *d* dimensions [Zhang et al, 2017].

So far, together these results faded the expectation to find tips to distinguish the models that generalize well from the models, which can only memorize training data, using models' *behavior during the training.* If these expectations would materialize, then they would shed light on the interpretability of the models too. We would be able, to filter out as unexplainable, the models that behave specifically for non-generalizable models.

However, the situation is different due to actual results. We can try to explain models that are accurate on training data using the heatmap activation method that identifies salient pixels. Obviously, we can compute these pixels and *"explain" complete noise*. To distinguish it from a meaningful explanation, we would need to use a traditional ML approach -- analyze errors beyond training data on the test data. Even this will not fully resolve the issue. It is commonly assumed that for the success of the ML model, training, validation and testing data should be from the *same probability population*. In the same way, the noise training, validation and testing data can be taken from a single population.

How to distinguish between the models trained on the true labels that are potentially explainable and the models trained on random labels that should not be meaningfully explainable? This is an open question for *the black box ML metho*ds.

The conceptual explanation methods based on the *domain knowledge* for the glass box ML models are equipped much better to solve this problem.

### 8.2. Visual Explanations for DNN

**Visualizing activations for texts**. The LSTMVis system [Kahng, et al. 2018] is for interactive exploration of the learnt behavior of hidden nodes in LSTM network. A user selects a phrase, e.g., "a little prince," and specifies a threshold. The system shows hidden nodes with activation values greater than the threshold and finds other phrases for which the same hidden nodes are highly activated. Given a phrase in a document, the line graphs and heatmap visualize the activation patterns of hidden nodes over the phrase.

Several other systems employ activation, heatmap, and parallel coordinates too. The open questions for all of them for model explanation are: why the activation should make sense for the user, (2) how to capture relations between salient elements, (3) how to measure that the explanation is right?

**Heatmap based methods for images**. Some alternative methods to find salient pixels in DNN include:

(i) sensitivity analysis by using partial derivatives of the activation function to find the max of its gradient,
(ii) Taylor decomposition of the activation functions by using its first-order components to find scores for the pixels,
(iii) Layer-wise relevance propagation (LRP) by mapping the activation value to the prior layers, and
(iv) blocking (occluding, perturbing) sets of pixels and finding sets, which cause the largest change of activation value that can be accompanies by the class change of the image [Montavon, et al, 2018].

More approaches are reviewed and compared in [Ancona et al, 2017; Choo, Liu, 2018; Gilpin et al, 2018, Zhang, Zhu, 2018].

From visualization viewpoint different methods that identify salient pixels in input images as explanation all are in the same category, because all of them use the same visualization heatmap method. The variations are that salient pixels can be shown in a separate image or as overlay/outline on the input image. The ways how they identify salient pixels are **black boxes**, for the end users. The only differences that users can observe is how well salient pixels *separate* objects of interest from the background (horses and bird in Fig. 31 [Montavon et al, 2018]), and how *specifically* they identify these objects (each horse is framed or not). The visualization of *features* that led to the conclusion typically needs other visualization tools, beyond the heatmap capabilities. Therefore, heatmap explanation is **incomplete explanation**.

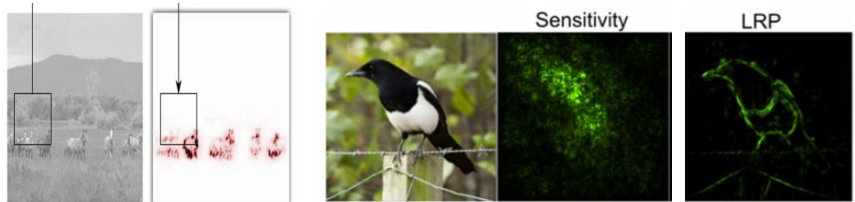

Fig. 31. DNN heatmap for classes "horse" and "bird" [Montavon et al, 2018; Samek et al, 2016].

**Visualizing intermediate layers**. While heatmaps overlaid on the original image to show salient pixels is a common way to explain DNN discoveries, heatmaps are also used to show features at the intermediate layers and to compare internal representations of object-centric and scene-centric networks [Zhou et al, 2014] that can be used for model explanation. However, it is even more difficult to represent in the domain terms, because in contrast with the input data/ image the layers are less connected to the domain concepts.

### 8.3. Rule-Based methods for Deep Learning

**RuleMatrix**. In [Ming et al, 2018] the rule-based explanatory representation is visualized in the form of RuleMatrix, where each row represents a rule, and each column is a feature used in the rules. These rules are found by approximating the ML model that was already discovered. As we see in Fig. 32, it is analogous to visualization of the association rules presented in section 3 (Fig.3). In Fig. 32, the rule quality measures are separate columns, while, in Fig. 3, they are integrated with the rules by using a heatmap approach. Next, Fig. 32 provides more information about each rule in the Matrix form, while Fig.3 provides more information in the form of parallel sets.

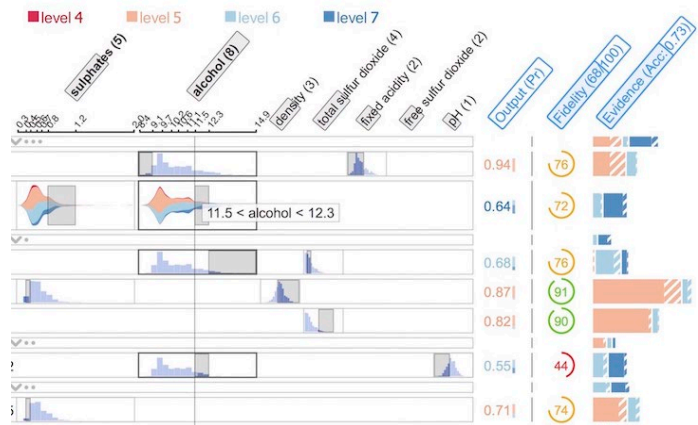

Fig. 32. Visualization of rules in Rule Matrix [Ming et al, 2018].

These authors point out that the current visualization tools focus on addressing the *needs of machine learning researchers* and developers, without much attention to *help domain experts* who have little or no knowledge of machine learning or deep learning [Ahmad et al 2018]. While it is true that domain experts have little ML knowledge, the issue is much deeper. All of us are domain "experts" in recognizing cats, dogs, birds, horses, boats and digits in the pictures. Consider the question that how many ML experts would agree that their knowledge of DNN and other ML algorithms allows them to say that they have an explanation how DNN recognized a cat vs. a dog? The meaningful explanation needs to be in the terms of features of cats and dogs, which is part of our commonsense knowledge. Similarly, for the domain experts, the meaningful explanation must be in their domain knowledge terms, not in the foreign ML terms.

**Interpreting Deep Learning Models via Decision Trees**. The idea of interpreting neural networks using decision trees can be traced to [Shavlik, 1996]. Now it is expanded to DNN to explain the prediction at the semantic level. In [Zhang et al, 2019], a decision tree decomposes feature representations into elementary concepts of object parts. The decision tree shows which object parts activate which filters, for the prediction, and how much each object part contributes to the prediction score. DNN is learned for object classification with disentangled representations, in the top conv-layer, where each filter represents a specific object part. The decision tree encodes various decision modes hidden inside the fully connected layers of the CNN in a coarse-to-fine manner. Given an input image, the decision tree infers a parse tree, to quantitatively analyze rationales for the CNN prediction, i.e. which object parts (or filters) are used for prediction, and how much an object part (or filter) contributes to the prediction

### 8.4. Human in the Loop Explanations

**Explanatory Interactive Learning**. DNN can use confounding factors within datasets to achieve high prediction of the trained models. These factors can be good predictors in a given dataset, but be useless in real world settings [Lapuschkin et al, 2019]. For instance, the model can be right in prediction, but for the wrong reasons, focusing incorrectly on areas outside of the issue of interest.

The available options include: (1) discarding such models and datasets, and (2) correcting such models by the human user interactively [Schramowski et al., 2020]. The corrections are penalizing decisions made for wrong reasons, adding more and better training cases including counterexamples, and annotated masks during the learning loop. While these authors report success in this explanation, a user cannot review thousands of images on correctness of heatmaps in training and validation data. This review process is not scalable to thousands of images.

**Explanatory graphs**. A promising approach to provide human-interpretable graphical representations of DNN are explanatory graphs [Zhang, Zhu, 2018] that allow representing the semantic hierarchy hidden inside a CNN. The explanatory graph has multiple layers. Each graph layer corresponds to a specific conv-layer of a CNN. Each filter in a conv-layer may represent the appearance of different object parts. Each input image can only trigger a small subset of part patterns (nodes) in the explanatory graph. The major challenge for the explanatory graphs is that they are derived from the DNN, if DNN is not rich enough to capture semantics, it cannot be derived. It follows from the fact that the explanation cannot be better, than its base model itself.

### 8.5. Understanding Generative Adversarial Networks GANs via Explanations

In GAN the generative network generates candidates while the *discriminative network* evaluates them, however, visualization and understanding of GANs is largely missing. A framework to visualize and understand GANs at the unit, object, and scene level proposed in [Bau et al, 2018] is illustrated by the following example. Consider images of the buildings without a visible door. A user inserts a door into each of them at the generative stage and then the discriminative network evaluates them. The abilities of this network to discover the door depends on the local context of the image. In this way, the context can be learned and used for explanation. The opposite common idea is covering some pixels to find the salient pixels.

## 9. Open Problems and Current Research Frontiers

### 9.1. Evaluation and development of new visual methods

An open problem for the visual methods intended to explain what a deep neural network has learned is **matching the** salient features discovered by explanation methods with human expertise and intuition. If, for a given problem, such matching is not feasible, then the explanation method is said to have failed the evaluation test. Existing explanation methods need to go through rigorous evaluation tests to be widely adoptable. Such methods are necessary as they can help in guiding the discovery of a new **explainable visualization** of what a deep neural network has learned.

The example below illustrates these challenges. The difference in explanation power between three heatmap visualizations for digit '3' in shown in [Samek et al, 2017]. See Fig. 33. The heatmap on the left is a randomly generated heatmap that *does not deliver interpretable information* relevant to '3'. The heatmap in the middle shows the whole digit *without relevant parts*, say, for distinguishing '3' from '8' or '9', but separate '3' from the background well. The heatmap of the right provides a *relevant information* for distinguishing between '3', '8' and '9'. If these salient pixels are provided by the ML classifier that discriminates these three digits then these salient pixels are consistent/**matched** with human intuition on differences between '3', '8' and '9'. However, for distinguishing '3' and '2', these pixels are not so salient for humans.

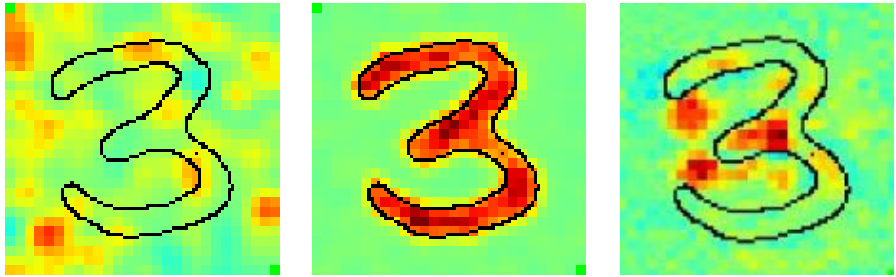

Fig 33. Difference between explainability of heatmaps [Samek et al, 2017]

### 9.2. Cross Domain Pollination: Physics & Domain based Methods

A promising area of potentially new insights for explainable methods is the intersection of machine learning with other well stablished disciplines like Physics [Ahmad et al 2020], Biology [Bongard,2009] etc. which have a history of explainable visual methods in their domains, e.g., the mathematical expressions describing the behavior and interaction of subatomic particles are quite complex, but they can be described via Feynman diagrams which a visual device for representing their interactions [Feynman, 1949]. Similarly physics inspired models are now being used to simplify and inform machine learning models which readily come up explainability [Ahmed et al 2020] but can still be limited by visualizing of large number of contributing variables. A promising direction would be to combine such physics based methods with GLC family of methods described above.

### 9.3. Cross-Domain Pollination: Heatmap for non-image data

Recent ML progress has been guided by cross-pollination of different subfields of ML and related computer science fields. This chapter illustrates multiple such examples of integration of machine learning, visualization and visual analytics. Deep neural networks algorithms have shown remarkable success in solving image recognition problems. Several DNN architectures developed for one type of images have been successful also in other types of images, demonstrating efficiency of knowledge transfer to other types of images.

Converting *non-image data to images* by using visualization expands this *knowledge transfer opportunity* to solve a wide variety of Machine Learning problems [Dovhalets et al, 2018, Sharma et al, 2019]: In such methods, a non-image classification problem is converted into the image recognition problem to be solved by powerful DNN algorithms. The example below is a combination of **CPC-R** and **CNN algorithms**. The CPC-R algorithm [Kovalerchuk et al, 2020] is converting non-image data to images, and the CNN algorithm discovers the classification model in these images. Each image represents a single numeric n-D point, as a set of cells with a different level of intensities and colors.

The CPC-R algorithm first splits attributes of an n-D point **x**=$(x_1,x_2,…,x_n)$ into consecutive pairs $(x_1,x_2)$, $(x_3,x_4)$, …$(x_{n-1},x_n)$. If *n* is an odd number, then the last attribute is repeated to get $n+1$ attributes. Then all pairs are shown as 2-D points in the same 2-D Cartesian coordinates.

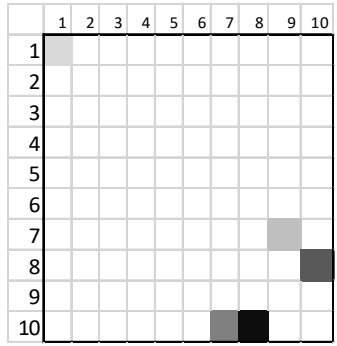 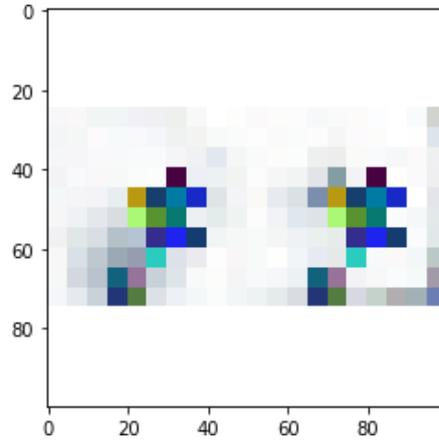

**(a)** 10-D point (8, 10, 10, 8, 7,10, 9,7,1,1) in CPC-R.

(b) Visualization in colored CPC-R of a case superimposed with mean images of two classes put side by side.

Fig. 34. CPC-R visualization of non-image 10-D points.

In Fig. 34, the CPC-R algorithm uses the grey scale intensity from black for $(x_1,x_2)$ and very light grey for $(x_{n-1},x_n)$. Alternatively, intensity of a color is used. This order of intensities allows full restoration of the order of the pairs from the image. In other words, a **heatmap** is created to represent each n-D point. The size of the cells can be varied from a single pixel to dozens of pixels. For instance, if each attribute has 10 different values then a small image with 10x10 pixels can represent a 10-D point by locating five grey scale pixels in this image. This visualization is lossless when values of all pairs $(x_i, x_{i+1})$ are different and do not repeat. An algorithm for treatment of colliding pairs is presented in [Kovalerchuk, et al, 2020].

Fig. 34(a) shows the basic CPC-R image design and Figure 34(b) shows a more complex design of images, where a colored CPC-R visualization of a case is superimposed with mean images of the two classes, which are put side by side, creating double images. The experiments with such images produce accuracy between 97.36% and 97.80% in 10-fold cross-validation for different CNN architectures for Wisconsin Breast Cancer data [Kovalerchuk, et al, 2020]. The advantage of CPC-R is in *lossless visualization* of n-D points. It also opens an *opportunity* discovering explanations in the form of *salient pixels/features* as it is done for DNN algorithms and described in the previous sections. In this way non-image data and ML models will get visual explanations.

## 9.4. Future Directions

Interpretable machine learning or explainable AI has been an active area of research for the last few years. Outside of a few notable exceptions, generalized visual methods for generating deep explanations in machine learning has not progressed as much given the centrality of visualization in human perceptual understanding. Future directions in interpretability research and applications are diverse and are informed by domain needs, technical challenges, ethical and legal constraints, cognitive limitations etc. Below is an incomplete list of some prominent challenges facing the field today:

- Creating simplified explainable models with prediction that humans can actually understand.
- "Downgrading" complex deep learning models for humans to understand them.
- Expanding visual and hybrid explanation models.
- Further developing of explainable Graph Models.
- Further developing of ML models in First Order Logic (FOL) terms of the domain ontology.
- Generating advanced models with the sole purpose of explanation.
- Post-training rule-extraction.
- Expert-in-the-loop in the training and testing stages with auditing models to check generalizability of models to wider real-world data.
- Rich semantic labeling of a model's features that the users can understand.
- Estimating the causal impact of a given feature on model prediction accuracy.
- Using new techniques such as counter-factual probes, generalized additive models, generative adversarial network technique for explanations.
- Further developing heatmap visual explanations of CNN by Gradient-weighted Class Activation Mapping and other methods with highlighting the salient image areas.
- Adding explainability to DNN architectures by layer-wise specificity of the targets at each layer.

## 10. Conclusion

Interpretability of machine learning models is a vast topic that has grown in prominence over the course of the last few years. The importance of visual methods for interpretability is being recognized as more and more limitations of real-world systems are coming into prominence. The chapter covered the motivations for interpretability, foundations of interpretability, discovering visual interpretable models, limits of visual interpretability in deep learning, a user-centric view of interpretability of visual methods, open problems and current research frontiers. The chapter demonstrated that the approaches for discovering the ML models aided by visual methods are diverse and expanding as new challenges emerge. This chapter surveyed current explainable machine learning approaches and studies toward deep explainable machine learning via visual means. The major current challenge in this field is that many explanations are still rather quasi-explanations and are often geared towards the ML experts rather than the domain user. There are often trade-offs required to create the models as explainable which requires loss of information, and thus loss of fidelity. This observation is also captured by theoretical limits, in regard to preserving n-D distances in lower dimensions presented based on the Johnson-Lindenstrauss Lemma for point-to-point approaches. The chapter also explored that additional studies, beyond the arbitrary points, explored in this lemma are needed for the point-to-point approaches.

Many of the limitations of the current quasi-explanations, and the loss of interpretable information can be contrasted with new methods like **point-to-graph GLC approaches** that do not suffer from the theoretical and practical limitations described in this chapter. The power of the GLC family of approaches was demonstrated via several real-world case studies, based on multiple GLC-based algorithms. The advantages of the GLC methods were shown, and suggestions for additional multiple enhancements was also discussed. The dimension reduction, classification and clustering methods described in this chapter support scalability and interpretability in a variety of settings: These methods include the visual PCA interpretation with GLC clustering for cutting the number of points etc. Lastly, we also discussed several methods that are used for interpreting traditional machine learning problem, e.g., visualization association rules via matrix and parallel set visualizations, data flow tracing for decision trees, visual analysis of Random Forests etc. PCA and t-SNE correspond to a class of models that are used for data and model understanding. While these methods are useful for a high-level data summarization, they also suffer from simplification, are lossy, and have distortion bias. The GLC based methods, being interpretable and lossless, however do not suffer from many of the limitations of these methods described here

Deep learning models have resisted yielding to methods that not only provide explainability but also have high model fidelity, mainly because of the model complexity inherent in deep learning models. A brief survey of interpretable methods in deep learning is also given in this chapter alo,ng with the strength and weakness of these methods. The need for explanations via heatmaps and for time series data is also covered. It is likely that heatmap implicit visual explanations will continue to be in the focus of further studies, while this chapter has shown the need to go beyond heatmaps in the future. What the examples demonstrate is that the landscape of interpretability is uneven in the sense that some domains have not been explored as compared to others. The human element in the machine learning system may prove to be the most crucial element in creating the interpretable methods. Lastly, it is noted that despite the fact that much progress has been made in interpretability of machine learning methods and the promise offered by visual methods, there is still a lot of work that needs to be done in this field to create systems that are auditable and safe. We hope that the visual methods outlined in this chapter will provide impetus for further development of this area, and help towards understanding and developing of new methods for multidimensional data across domains.